\documentclass[lettersize,journal]{IEEEtran}
\usepackage{amsmath,amsfonts}
\usepackage{array}
\usepackage[caption=false,font=footnotesize,labelfont=rm,textfont=rm]{subfig}
\usepackage{textcomp}
\usepackage{color}
\usepackage{stfloats}
\usepackage{url}
\usepackage{verbatim}
\usepackage{graphicx}
\usepackage{cite}
\usepackage{booktabs}
\usepackage{hyperref}
\usepackage{multirow}
\usepackage{makecell}
\usepackage{epsfig}
\usepackage{amsfonts}
\usepackage{amssymb}
\usepackage{caption}
\usepackage{footnote}
\graphicspath{{figures/}}

\makeatletter
\newcommand\figcaption{\def\@captype{figure}\caption}
\newcommand\tabcaption{\def\@captype{table}\caption}
\makeatother

\usepackage{algorithm}
\usepackage{algorithmicx}
\usepackage{algpseudocode}

\begin{document}

\title{Fine-Grained Domain Generalization 

with Feature Structuralization}

\author{Wenlong Yu,
        Dongyue Chen,
        Qilong Wang,~\IEEEmembership{Member,~IEEE}
        Qinghua Hu,~\IEEEmembership{Senior~Member,~IEEE}
        
\IEEEcompsocitemizethanks{\IEEEcompsocthanksitem 
This work was partly supported by the National Natural Science Foundation of China under Grants 61925602, U23B2049, 62406219, and 62436001; and by the China Postdoctoral Science Foundation-Tianjin Joint Support Program under Grant 2023T014TJ.
\emph{(Corresponding author: Dongyue Chen)}.


Wenlong Yu, Dongyue Chen, Qilong Wang, and Qinghua Hu are with the College of Intelligence and Computing, Tianjin University, Tianjin 300350, China, and also with the Tianjin Key Lab of Machine Learning, Tianjin University, Tianjin 300350, China (e-mail: wlong\_yu@126.com, dyue\_chen@163.com, qlwang@tju.edu.cn, huqinghua@tju.edu.cn). 


}}


\markboth{IEEE Transactions on Multimedia}%
{Shell \MakeLowercase{\textit{et al.}}: A Sample Article Using IEEEtran.cls for IEEE Journals}


\maketitle


\begin{abstract}
Fine-grained domain generalization (FGDG) is a more challenging task than traditional DG tasks due to its small inter-class variations and relatively large intra-class disparities. 
When domain distribution changes, the vulnerability of subtle features leads to a severe deterioration in model performance.
Nevertheless, humans inherently demonstrate the capacity for generalizing to out-of-distribution data, leveraging structured multi-granularity knowledge that emerges from discerning the commonality and specificity within categories.
Likewise, we propose a Feature Structuralized Domain Generalization (FSDG) model, wherein features experience structuralization into common, specific, and confounding segments, harmoniously aligned with their relevant semantic concepts, to elevate performance in FGDG. 
Specifically, feature structuralization (FS) is accomplished through joint optimization of five constraints: a decorrelation function applied to disentangled segments, three constraints ensuring common feature consistency and specific feature distinctiveness, and a prediction calibration term.
By imposing these stipulations, FSDG is prompted to disentangle and align features based on multi-granularity knowledge, facilitating robust subtle distinctions among categories.
Extensive experimentation on three benchmarks consistently validates the superiority of FSDG over state-of-the-art counterparts, with an average improvement of 6.2\% in FGDG performance. 
Beyond that, the explainability analysis on explicit concept matching intensity between the shared concepts among categories and the model channels, along with experiments on various mainstream model architectures, substantiates the validity of FS.

\end{abstract}


\section{Introduction}
\IEEEPARstart{D}{eep} Learning (DL) has achieved remarkable success in various visual tasks thanks to its powerful ability to learn and extract representations from data.
However, most data-driven models, operating under the oversimplified assumption of independent and identically distributed (i.i.d.) scenarios, encounter obstacles when deployed in diverse contexts, often referred to as out-of-distribution (OOD) problems.
The recognition capability of deep neural networks (DNNs), trained on a source domain (e.g., sketches), significantly deteriorates upon application to other target domains (e.g., the real world) ~\cite{dgtmm1}. 
To address OOD problems, many generalization-related topics have been proposed, such as domain adaptation (DA) and domain generalization (DG). 
Among them, DG is a more credible setting since it does not incorporate test domain data into the training process \cite{dgsurveybaidu}.
Typically, identifying domain invariance to reduce the risk of overfitting is a viable direction of DG \cite{riskirm, causal1}. 
It treats the invariance across given source domains as intrinsic generalizable features of objects \cite{dgsurvey}.

\begin{figure}[t]
\begin{center}
\centerline{\includegraphics[width=\columnwidth]{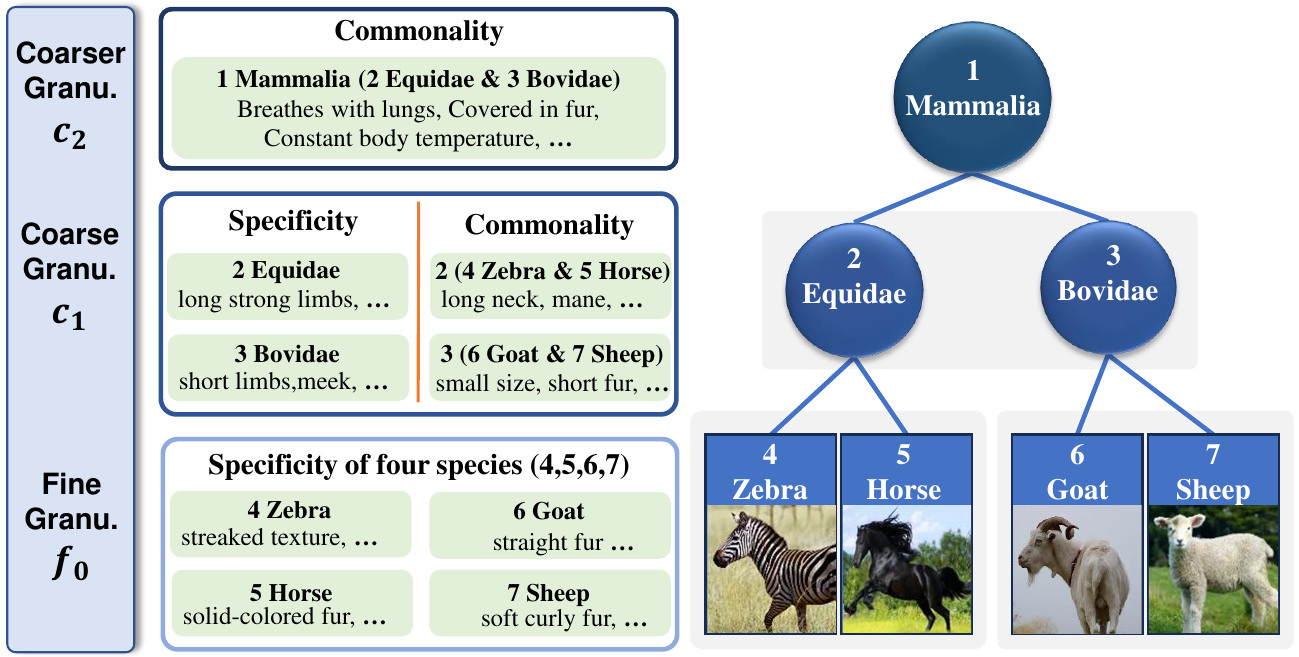}}
\caption{Instance of multi-granularity knowledge. Four animals are categorized into various classes across three granularity levels according to their commonalities and specificities, as described in the green region.
The numbers represent species labels: labels 4, 5, 6, and 7 correspond to the $f_0$ granularity level; 2 and 3 to the $c_1$ level; and 1 to the coarsest $c_2$ level. 
For commonalities, the number outside parentheses indicates the parent category, while the two inside are its sub-categories.
}
\label{shuxingeg}
\end{center}
\vspace{-0.5in}
\end{figure}

However, this invariance performs sub-optimally in fine-grained domain generalization tasks (FGDG), especially in single-source scenarios.
In fine-grained data, the distinctions among categories are relatively small compared to intra-class differences.
DNNs tend to learn finer discriminative features when trained on fine-grained data, resulting in a worse entanglement among spurious features and object categories \cite{finesurvey}.
The OOD problems are exacerbated as the learned features grow increasingly vulnerable and fragile to distribution shifts \cite{PAN}.
Furthermore, collecting multi-source fine-grained data is more laborious than collecting multi-source coarse-grained data \cite{robustfglabel}, and the reduced distributional diversity within a single-source training environment makes learning invariance more arduous \cite{dgsurvey}.
All of these factors significantly increase the difficulty of FGDG, resulting in traditional DG methods exhibiting subpar generalization performance.
Therefore, it is crucial to explore the fine-grained invariant representation capabilities of DNNs more thoroughly.

In FGDG tasks, fine-grained categories exhibit multi-granularity structures built upon the commonalities and specificities between categories.
This aspect is overlooked by previous DG methods. 
In contrast, humans exhibit a higher level of generalizability by leveraging these structured semantics for learning and recognizing.
According to information decomposition theory of the brain \cite{cellinformationdecompose}, family resemblances \cite{familyresemblances}, feature integration theory \cite{featureintegrationtheory}, and lexical structure \cite{conceptstruinbrain} in cognitive psychology, when learning to recognize objects,
people focus primarily on the salient targets and inductively disentangle the commonalities and specificities among them. 
Commonalities represent basic visual concepts, whereas specificities correspond to individual conceptual extensions. 
Both serve as the foundational elements for object classification and vary across distinct granularity levels \cite{panyunhe,baidusun,panyunhe2024}.
For instance, as Fig. \ref{shuxingeg} shows, four animals can be reclassified into categories of Equidae and Bovidae based on their commonalities and specificities from a higher $c_1$ granularity level, respectively. 
At $c_1$ level, zebra and horse share certain characteristics, such as long necks, while manifesting specificities at fine-grained level $f_0$. 
It can be observed that the common and specific features, which constitute the intrinsic characteristics of objects, can be facilitated by incorporating multi-granularity knowledge.
Understanding the commonalities and specificities between categories enhances the generalizability, as both focus more on the intrinsic features of the objects themselves.

However, a key challenge arises: how to explicitly embed these structured commonalities and specificities into DL models. 
In particular, it is unclear to determine what semantic basis can be used for constructing clusters of commonalities and specificities, and it is challenging to constrain them within their respective semantic clusters, as literature has discussed the unattainability of learning decoupling semantics in the absence of sufficient additional knowledge constraints \cite{icml2019best}.

To address this challenge, we propose a Feature Structuralization (FS) framework employing multi-granularity knowledge as an additional constraint for semantic disentanglement and alignment.
Researches in the explainability field show that features or tokens behave as distinct semantics \cite{networkdessection,whatistabby,naturecrp,clipx}, which implies that commonalities and specificities can be disentangled from the learned features.
Acknowledging that recognition tasks predominantly seek to discern target objects in an image that encompasses target semantics and confounding, we disentangle the total semantical features learned from an image into three components, namely commonality, specificity, and confounding. 
Furthermore, the representations of commonalities and specificities are constrained within their respective semantic clusters with the aid of a multi-granularity knowledge structure.
The collaboration between both facilitates the network in learning category invariance while preserving learning discrimination.
Consequently, FS optimizes the fine-grained invariant representations as both focus more on the attributes of the objects themselves rather than on the other spurious correlations.


To formulate FS, we devise a pipeline consisting of two essential steps: disentanglement and alignment.
We disentangle the three parts according to their channel indices. 
A decorrelation optimization function is employed to minimize the mutual information between them. 
It enables the confounding segment to learn complementary and redundant information for recognition \cite{nccausaldecouple, random_in_generali,redundant_help_generali}.
Three similarity-based losses are then proposed for the alignment of common and specific parts with the aid of multi-granularity knowledge. 
Two of these are leveraged to pull commonalities closer across diverse granularities, while the third serves to create distance among the specific components.
Furthermore, a Feature Structuralized Domain Generalization model (FSDG) is constructed by incorporating a prediction calibration method. 
Three model variations are expanded in conjunction with Convolutional Neural Network (CNN), Transformer, and Multi-Layer Perceptron (MLP).

In summary, the contributions of this work are as follows:

(1) We propose an FS framework to address a more difficult FGDG challenge, drawing inspiration from cognitive psychology. 
This novel approach integrates data and multi-granularity knowledge to structure the learned features into common, specific, and confounding parts according to their channel indices, which is in line with the analysis of explainability. 
FS endows the model with the ability to explicitly learn commonalities and specificities between categories.

(2) We construct a feature-structuralized model FSDG by formulating FS into a disentanglement and alignment pipeline. 
A disentanglement enhancement function is derived to reduce the correlation between the three segments.
Three constraints are then designed to achieve semantic alignment for the common and specific components. 
To our knowledge, FSDG is a pioneer tailored to address FGDG issues.

(3) A CRP method is refined to substantiate the validity of FS. 
It statistically computes the most relevant concepts for every category, showing that FS significantly increases the explicit concept matching intensity between the shared concepts among categories and the model channels. 
Besides, extensive experiments on three benchmarks, along with three FSDG variants, show that FSDGs outperform their counterparts in terms of FGDG performance.

The remainder of this article is organized as follows. 
Section II reviews related works. 
Problem formulation and the proposed FS framework are presented in Section III. 
In Section IV, the experimental analysis of the proposed approach is presented. 
The explainability analysis is explored in Section V. Finally, Section VI concludes this article with a brief discussion on limitations and future work.

\begin{figure*}[t]
\begin{center} 
    \centerline{\includegraphics[width=7.1in]{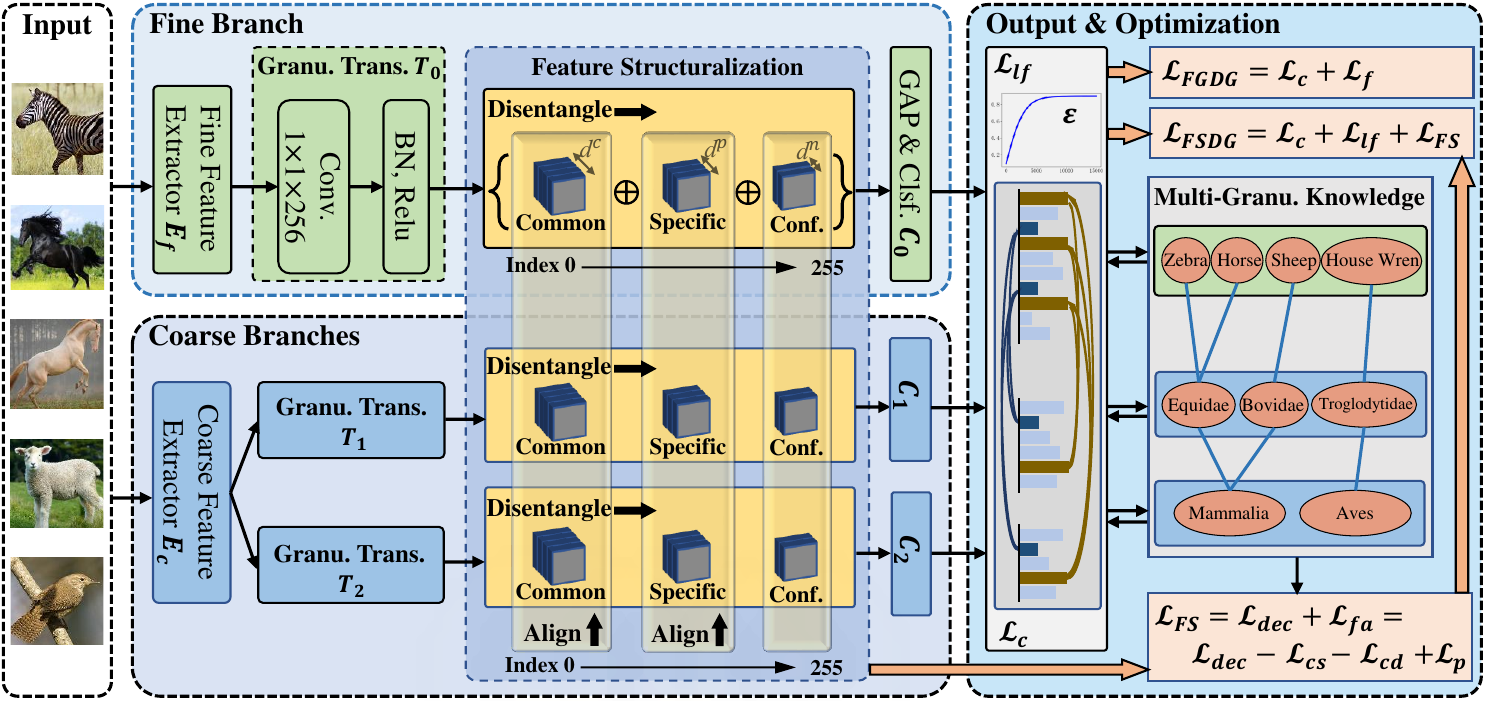}}
\caption{Illustration of the model exemplified in a three-granularity hierarchy (i.e., $G=3$). The FS module is highlighted in the middle box with solid arrows depicting the operation dimensions. $\oplus$ represents the disentanglement operator. 
Given five input images, FSDG outputs multi-granular results, as indicated in the right section. 
The box showing $\mathcal{L}_{lf}$ illustrates the alignment operation for coarse-fine predicted distributions and shows $\varepsilon$ during the training process. 
Conf. is an abbreviation of confounding and Granu. Trans. means the Granularity Transition Layer.
The coarse branches and the FS optimization module are excluded during model inference.}
\label{model}
\end{center}
\vspace{-0.3in}
\end{figure*}

\section{Related Work}  
\label{relatedwork}

\subsection{Domain Generalization}
Research on alleviating OOD issues has been conducted from various perspectives \cite{dgsurveybaidu,dgtmm3}.   
DA, as a straightforward solution, directly utilizes target domain data as auxiliary knowledge to train the model \cite{PAN}.   
However, collecting or even identifying target data before deploying the model is unbearable, and when applied to a third domain, the model still encounters failures \cite{dgsurvey}.   
Therefore, it is more appropriate to study the DG problem without training the target domain data.
In recent years, numerous DG strategies have been proposed, such as 
learning domain-invariant representations \cite{mmd,coral}, 
data augmentation \cite{mixstyle,mixup}, 
optimization strategies \cite{sagm,arm,mldg}, and
disentangled representations \cite{disentangle2,dgtmm2}.

Robust domain-invariant representations emphasize computing the consistencies across multi-source domains \cite{dgsurvey}. 
Rosenfeld et al. \cite{riskirm} acquired domain-invariant features with the help of causal analysis. 
Literature \cite{ridg} computed a kind of rationale invariance to enhance DG performance. 
Data augmentation methods can be used to generate multiple domain data for single-source DG problems \cite{dataaugsingle}. 
Zhou et al. \cite{mixstyle} mixed the statistics of two images in the image or feature level to generate new styles.
Wang et al. \cite{sagm} improved DG capability by implicitly aligning the gradient directions between the empirical risk and the perturbed loss.
DualVAE \cite{dgtmm4} utilized a dual-path VAE architecture to disentangle latent factors and trained a classifier on top of the semantic factor for DG issues.
Low-rank decomposition on weight matrices is applied in \cite{disentangle4} to identify features that are more generalizable.

However, a majority of these methods are grounded in multi-source data, without consideration of learning invariance through understanding commonalities and specificities between categories. 
They struggle to achieve the anticipated performance in single-source and FGDG tasks.

\subsection{Fine-Grained Visual Categorization}

Fine-grained visual categorization (FGVC) strives to discriminate subtle distinctions between subordinate categories within the same root category \cite{finesurvey}. 
It presents a formidable challenge due to the inherent dilemma of minimal inter-class variation coupled with significant intra-class variation.
Researchers endeavoring to handle this problem can be classified into three main paradigms: feature enhancement-based \cite{fgvc2}, localization-based \cite{fgvclocal1,fvtmm1}, and external knowledge-based methods \cite{sun2022sim}. 
These methods mainly focus on enhancing the fine discriminative characteristics.
For example, FET-FGVC \cite{fgvc2} proposed a feature-enhanced Transformer, cooperating with Graph Neural Networks, to improve the performance of FGVC.
Zheng et al. \cite{fgvclocal1} firstly localized key object parts within images and then found discriminative clues on object parts.
Literature \cite{sun2022sim} incorporated object structure knowledge into the transformer to upgrade discriminative representations.
As for fine-grained generalization issues, PAN \cite{PAN} harnessed multi-granularity knowledge, aligning the final multi-granular predictions to improve fine-grained DA performance. 
WDAN \cite{wdan} utilized a weighted discriminative adversarial network to release the fine-grained DA issue.
Besides, LDRE \cite{ldre} utilized VLMs to retrieve a target image based on the commonalities between the reference and candidates.
However, these methods neglect feature invariance between categories, and the single-source FGDG problem has not been explored.

Compared with all aforementioned methods, FS aims to tackle the problem of single-source FGDG. 
Our method can be regarded as a cognition-inspired paradigm \cite{conceptstruinbrain,brainstructure} in which FS organizes the feature space into commonality, specificity, and confounding through multi-granularity knowledge.

\section{Method}
\label{3method}

In this section, we introduce our problem formulation and the details of our proposed FS framework. We illustrate how our method disentangles feature space into three parts. An aligning strategy is then adopted to achieve semantic functionalization.

\subsection{Problem Statement}
\label{pro_state}

\textbf{Domain Generalization:}
Let $\mathcal{X}=\left\{(x)\right\}$ be the input space and $\mathcal{Y}=\left\{(y)\right\}$ the label space.
In the context of DG, 
$\mathcal{D_S} = \left\{\left(\mathcal{X}^s, \mathcal{Y}^s\right)\right\}_{s=1}^{M}$ 
and $\mathcal{D_T} = \left\{\mathcal{T}\right\}$ denote $M$ source domains and an unseen target domain, respectively.
We train a model $f: \mathcal{X} \rightarrow \mathcal{Y}$ utilizing only source domain data $\mathcal{D_S}$.
The goal of DG is to minimize the prediction error of $f$ on the unseen target domain $\mathcal{D_T}$.
Classical DG approaches work in multi-source scenarios (i.e., $M>1$), learning invariant representations of multiple source domains. 
In contrast, single-source DG (i.e., $M=1$) is a more realistic problem in open environments, as it is hard to ensure that invariant features remain consistent in the target domain, especially given the limited number of available source domains for training.

\textbf{Fine-Grained Domain Generalization:}
The label space of $\mathcal{D_S}$ is constructed in a multi-granular format with a total $G$-layer hierarchy (i.e., $\mathcal{Y}=\left\{(y_f,\left.y_g\right|_{g=1}^{G-1})\right\}$).
Traditional DG can be seen as a particular kind of FGDG whose label space has only the fine-grained labels $y_f$ without the coarse-grained labels $\left.y_g\right|_{g=1} ^{G-1}$, where $g$ is a notation of granularity level and subscript $f$ represents the fine-grained content (i.e., $g=0$).
The objective of FGDG is to minimize the fine-grained prediction error on the unseen target domain $\mathcal{D_T}$.
In this paper, we enhance FGDG performance in a single-source scenario (i.e., $M=1$), taking into account a more realistic situation in which multi-source fine-grained structured data is more difficult to fabricate.  
Additionally, single-source FGDG can demonstrate the efficacy of our method more effectively.

Accordingly, we construct a baseline FGDG model as shown in Fig. \ref{model}. 
Macroscopically, feature extractors $E$ extract semantic features, followed by granularity transition layers $T$ to filter the corresponding granularity information. 
Finally, the Global Average Pooling (GAP) and Fully Connected (FC) classifier $C$ predict the resulting distributions. 
Integrating disentanglement and alignment, the FS optimization module is introduced to structuralize the features systematically. 

\subsection{Disentanglement and Decorrelation of Three Partitions}
\label{3.1disent}

In this paper, our objective is to enhance FGDG performance utilizing multi-granularity knowledge. 
In order to fully exploit this knowledge, we configure two backbones to extract coarse and fine features separately, as shown in Fig. \ref{model}. 
$G$ transition layers parallelly decompose the coarse and fine backbone features into features unique to each granularity, resulting in $G$ branches working in a multi-task manner. 
The fine branch is dedicated to identifying the fine-grained categories, while the features of coarse branches assist the fine-grained features in capturing commonalities and specificities.

Specifically, given a batch of source domain images 
$x \in \mathbb{R}^{B \times 3 \times W \times H}$, 
two feature extractors $E_c$ and $E_f$ extract coarse and fine features $\mathcal{F}_c$ and $\mathcal{F}_f$, respectively. 
Then $\left.T_g\right|_{g=0} ^{G-1}$ take them as input to decompose the granular features $\left.\mathcal{F}_g\right|_{g=0} ^{G-1}$ for every granularity.
For branch $g$, $\mathcal{F}_g$ consists of a set of features $\left\{{f}_{g}^{1}, {f}_{g}^{2}, \cdots,{f}_{g}^{d} \right\}$, where $d$ is the channel number of $T_g$. 
Finally, $\left.C_g\right|_{g=0} ^{G-1}$ output the predicted logits belonging to every granularity. 
It is imperative to elucidate that the auxiliary coarse branches and the entire FS optimization module will be excluded during inference, signifying that the ultimate deployed model comprises exclusively the fine-grained branch. 
Moreover, the feasibility of training with a solitary backbone is also evident.

Features ${f}_{g} \in \mathbb{R}^{B \times 1 \times w \times h}$ of CNN or tokens of Transformer represent distinctive semantics of the input images ~\cite{naturecrp,clipx}. 
Taking CNN as an example, we obtain $d$ semantic features $\mathcal{F}_g$ after $T_g$. We disentangle them into three segments: 
\begin{equation} 
\label{f_disentangle}
    \left\{\mathcal{F}_g^{c}, \mathcal{F}_g^{p}, \mathcal{F}_g^{n}\right\} = Disentangle(\mathcal{F}_g).
\end{equation}
In this paper, we disentangle features at the conceptual level according to their channel indices, as the interpretability analysis shows that channels can express various concepts \cite{naturecrp}. 
This procedure is also characterized by its conciseness, practicality, and computational efficiency. 
It allows us to explicitly identify the index positions of the functionality of the three parts, fortifying the transparency of the model. 
Therefore, $\mathcal{F}_g^{c}=\left\{{f}_{g}^{1}, \cdots, {f}_{g}^{{d^{c}}}\right\}$ has $d^{c}$ common features, 
$\mathcal{F}_g^{p}=\left\{{f}_{g}^{d^{c}+1}, \cdots, {f}_{g}^{{d^{c}}+{d^{p}}}\right\}$ reflects the ${d^{p}}$ specific features,
$\mathcal{F}_g^{n}=\left\{{f}_{g}^{d^{c}+{d^{p}}+1}, \cdots, {f}_{g}^{{d}}\right\}$ is the confounding segment, where $ d = d^{c} + d^{p} + d^{n}$, the superscript of feature $f$ denotes its index, and the subscript of $f$ indicates the granularity level to which it belongs, respectively. 

To enhance the degree of disentanglement from a semantic perspective, we argue that the conceptual features described in the three segments should be as independent and orthogonal as possible.
We first compute the prototypes of the segments:
\begin{equation}
\label{proto_disentangle}
  \left\{
    \begin{gathered}
        P_{g}^c =  MEAN(F_{g}^{c}, dim=1)
        \\
        P_{g}^p =  MEAN(F_{g}^{p}, dim=1)
        \\ 
        P_{g}^n =  MEAN(F_{g}^{n}, dim=1),
    \end{gathered}
  \right.
\end{equation}
where $MEAN(X, dim)$ is the average operator of tensor $X$ along the $dim$ dimension, $F_g^c \in \mathbb{R}^{B \times d^c \times wh}$, $F_g^p\in \mathbb{R}^{B \times d^p \times wh}$ and $F_g^n\in \mathbb{R}^{B \times d^n \times wh}$ are slices of the stacked 3-d tensor $F_g \in \mathbb{R}^{B \times d \times wh}$ along the channel dimension, with extraction ratios set to $r^c$, $r^p$ and $r^n$. 
The resulted $P_{g}^c \in \mathbb{R}^{B \times wh}$, $P_{g}^p$ and $P_{g}^n$ are then stacked as $P_{g}^{all} \in \mathbb{R}^{B \times 3 \times wh}$.
To effectively harness the parallel computing capabilities of GPUs, we compute the covariance of $P_{b,g}^{all} \in \mathbb{R}^{3 \times wh}$ of each sample $b$ in the batch $B$ as the distance between the 3 segments:
\begin{equation}
\label{l_dec}
    \mathcal{L}_{dec} = AVG(
    {\left\langle P_{b,g}^{all}, P_{b,g}^{all}\right\rangle 
    \over \Vert P_{b,g}^{all} \Vert_2 \Vert P_{b,g}^{all} \Vert_2} - \mathbf{I}).
\end{equation}
Cosine similarity is utilized as the metric because it directly quantifies orthogonality. 
Other metrics, such as Hilbert-Schmidt independence criterion (HSIC) \cite{hsic} and Euclidean distance \cite{euc}, are also tested, and results are presented in Section \ref{4experiments}.
Considering that the prototype vectors exhibit unity correlation with themselves (the diagonal elements), we subtract an identity matrix $\mathbf{I}$.
The elements of the covariance matrix, which subtracts self-correlation, yield the inter-correlation between the prototypes.
The objective loss function of the decorrelation is averaged (i.e. the $AVG(\cdot)$ operator) among all samples and granularities (subscripts $b$ and $g$).

\subsection{Alignment for Commonality, Specificity and Prediction}
\label{3.1framework}

In essence, our ambition is to endow each feature with a semantic concept that can describe the common and specific parts between objects. 
In the real world, commonalities and specificities can be identified and utilized to distinguish different objects with the assistance of multi-granularity knowledge.
In the same way, this multi-granular structure can effectively assist in identifying and aligning the learned features with those functions for DL models.

\textbf{Commonality Alignment} constrains the first section of features to match the common characteristics between categories. 
The constraint is implemented from two perspectives. 
Firstly, we argue that the common features within a sample should consistently manifest across diverse granularities because the intrinsic information regarding the objects remains unaffected by alterations in granularities. 
Secondly, we mandate the congruence of common features belonging to identical categories at the parent granularity but differing at the sub-granularity. 
For example, zebras and horses at the sub-granularity emerge as distinct sub-categories of the parent category Equidae while sharing certain common attributes, as shown in Fig. \ref{shuxingeg}.

In particular, the first constraint, denoted as $\mathcal{L}_{cs}$, measures how similar the common features of the same samples are across each pair of neighboring granularities. This distance is formulated as
\begin{equation} 
\label{ssdgc}
    S_{cs} =  AVG({
    \left\langle{f}_{b,g}^{c},{f}_{b,g+1}^{c}\right\rangle 
    \over \Vert {f}_{b,g}^{c} \Vert_2 \Vert {f}_{b,g+1}^{c} \Vert_2}
    ),
\end{equation}
where ${f}_{b,g}^{c} \in \mathbb{R}^{d^c \times wh}$ represents the common feature tensor belongs to ${g}_{th}$ granularity of ${b}_{th}$ image.
This distance is averaged across all samples and all parent granularities (amounting to a total of $G-1$ levels).


For the second constraint, we locate sub-granularity samples of every parent class, progressing from the lowest parent granularity to the highest. 
Given a batch of images $x$, assume there are a total of $Q_{g+1}$ classes (parent classes) at parent granularity level $g+1$. 
For the $q_{th}$ parent class, it comprises a total of $K_{q,g}$ subcategories at the sub-granularity level $g$. 
For each subcategory $k$, we locate all samples belonging to this subcategory, which are denoted as $B_{k,q,g}$. 
Accordingly, we derive the feature tensor $F_{B_{k,q,g}}\in \mathbb{R}^{B_{k,q,g} \times d \times wh}$ for these samples.
We then compute the GAP of the common segment $F_{B_{k,q,g}}^{c} \in \mathbb{R}^{B_{k,q,g} \times d^c \times wh}$ of the feature tensor to represent the fundamental common sub-centroid:
\begin{equation} 
\label{proto_dsdgc}
    P_{k,q,g}^c =  MEAN(GAP(F_{B_{k,q,g}}^{c}), dim=0).
\end{equation}
$P_{k,q,g}^c \in \mathbb{R}^{d^c}$ represents the centroid of the common characteristics of the current sub-category $k$ belonging to the parent category $q$.
The second constraint, denoted as $\mathcal{L}_{cd}$, 
forces the sub-centroids of a given parent category to move closer. 
Similar to Eq. \ref{l_dec}, we compute the covariance matrix of the stacked sub-centroid tensor $P_{q,g}^c \in \mathbb{R}^{K_{q,g} \times d^c}$ and subtract the self-correlation of each sub-centroid:
\begin{equation} 
\label{dsdgc}
    S_{cd} =
    AVG({\left\langle P_{q,g}^c, P_{q,g}^c\right\rangle 
    \over \Vert P_{q,g}^c \Vert_2 \Vert P_{q,g}^c \Vert_2} - \mathbf{I}
    ).
\end{equation}
The distance is also averaged across all parent categories and all granularities.

\textbf{Specificity Alignment} regulates the second group of features to match the distinctive characteristics between categories, denoted as $\mathcal{L}_{p}$. 
The main idea is that the specific representations of different categories at the same granularity should be maximally distinct, learning the differences between categories. 
Similarly, we locate all samples that belong to category $k$ at granularity $g$ and derive their specific feature tensor $F_{B_{k,g}}^p \in \mathbb{R}^{B_{k,g} \times d^p \times wh}$. The specificity centroid is defined as
\begin{equation} 
\label{proto_dssgp}
    P_{k,g}^p=MEAN\left(GAP\left(F_{B_{k,g}}^p\right),dim=0\right).
\end{equation}

The averaged distance among the stacked centroids $P_{g}^p \in \mathbb{R}^{K_g \times d^p}$ for all granularities can be formulated as
\begin{equation} 
\label{dssgp}
    S_{p} =  AVG({{\left\langle P_{g}^p, P_{g}^p\right\rangle 
    \over \Vert P_{g}^p \Vert_2 \Vert P_{g}^p \Vert_2} - \mathbf{I}}
    ).
\end{equation}

After these disentanglement, decorrelation, and alignment operations, the three partitions focus on recognizing different semantics. 
It is worth noting that no constraints are imposed on the confounding segment.
This segment maintains the model's redundancy and randomness, given the class-related nature of the first two partitions. 
The $\mathcal{L}_{dec}$ ensures that the third partition learns complementary information for recognition.

\textbf{Prediction Calibration} constrains the predictions from each branch in the FSDG model to construct a granularity tree. 
Regarding ~\cite{PAN,multilabel}, the fine-grained ground-truth labels progressively integrate with the predicted distributions at the coarse-grained branches, as shown in Fig. \ref{model}. 
The prediction alignment loss $\mathcal{L}_{lf}$ is designed to connect the prediction spaces of fine and coarse granularities during the training process and to optimize the fine branch. $\mathcal{L}_{lf}$ is formulated as
\begin{equation} 
\label{llh}
\begin{aligned}
    \mathcal{L}_{lf} (\left.\widehat{y}_g  \right| &  _{g=1}^{G-1}, \widehat{y}_f, \left.  y_f\right) \\ &
    = D_{\mathrm{KL}}\left(\varepsilon y_f +
    (1-\varepsilon) \sum_{g=1}^{G-1} \frac{\widehat{y}_g}{G-1} \| \widehat{y}_f\right),
\end{aligned}
\end{equation}
where $\widehat{y}_f = {C_f}({T_f}({E_f}(x)))$ is the fine-grained predicted distribution, $D_{\mathrm{KL}}$ is the Kullback-Leibler divergence, $\widehat{y}_g = {C_g}({T_g}({E_c}(x)))$ ($g\neq0$), of which dimension has been extended to the same as $\widehat{y}_f$, is the output of coarse branches, 
and $\varepsilon$
controls the intensity of the coarse-grained influence on fine-grained classification. 
All coarse branches are trained to minimize the traditional recognition objective:
\begin{equation}
\label{l_c}
    \mathcal{L}_{c} = \sum_{g=1}^{G-1} L_{CE}\left(\widehat{y}_g, y_g\right),
\end{equation}
where $L_{CE}$ represents the cross-entropy (CE) loss.

The FS is achieved by the FS objective function $\mathcal{L}_{FS}$, comprising disentanglement and alignment functions $\mathcal{L}_{dec}$ and $\mathcal{L}_{fa}$, and a prediction calibration loss $\mathcal{L}_{lf}$. $\mathcal{L}_{fa}$ maximizes the similarity of common features while minimizing the similarity of specific features. The FS objective is formulated as
\begin{equation} \label{lfs}
\begin{aligned}
    \mathcal{L}_{FS} & = \mathcal{L}_{dec} + \mathcal{L}_{fa}\\ &
    = \mathcal{L}_{dec} - \mathcal{L}_{cs} - \mathcal{L}_{cd} + \mathcal{L}_{p}\\
    &
    = \mathcal{L}_{dec} - \lambda_{cs}\mathcal{S}_{cs} - \lambda_{cd}\mathcal{S}_{cd} + \lambda_{p}\mathcal{S}_{p},
\end{aligned}
\end{equation}
where $\lambda_{cs}$, $\lambda_{cd}$, and $\lambda_{p}$ are coefficients for each loss.

Overall, the training loss of the proposed FSDG is
\begin{equation} \label{zongloss}
    \mathcal{L}_{FSDG} = \mathcal{L}_{c} + \mathcal{L}_{lf} + \mathcal{L}_{FS}.
\end{equation}

The model architecture of the baseline FGDG model is the same as that of FSDG.
The difference lies in the absence of $\mathcal{L}_{FS}$. Instead, the baseline FGDG model progressively optimizes the fine branch using $\mathcal{L}_{f}$ with the help of $\varepsilon$.
The loss function is formulated as
\begin{equation} \label{zonglossfgdg}
\begin{aligned}
    \mathcal{L}_{FGDG} & = \mathcal{L}_{c} + \mathcal{L}_{f}
    = \mathcal{L}_{c} + D_{\mathrm{KL}}\left(\varepsilon y_f \| \widehat{y}_f\right).
\end{aligned}
\end{equation}

\begin{algorithm*}[t]
\caption{Pseudocode of the FSDG Approach} 
\label{alg_1}
\begin{algorithmic}[1]
\Require
    Source-domain training samples $x^s$;
    Multi-granularity knowledge $\left(y_f^s,\left.y_g^s\right|_{g=1}^{G-1}\right)$; 
    coarse feature extractor $E_c$; 
    fine feature extractor $E_f$; 
    number of layers of the multi-granularity structure $G$;
    multi-granularity transition layers $\left.T_g\right|_{g=0} ^{G-1}$; 
    multi-granularity classifier $\left.C_g\right|_{g=0} ^{G-1}$; 
    channel dimensions $d$; 
    extraction ratios $r^c$, $r^p$ and $r^n$.
\vspace{0.05in}
\Ensure
    Well-trained fine-branch networks $E_f^\ast$, $T_0$ and $C_0$;
    \For{iteration = 1 : $N$}
        \State Compute granular features $\left.\mathcal{F}_g\right|_{g=0} ^{G-1}$ of $G$ transition layers;
        \For {g = 0 : $G-1$}
            \State Disentangle $\mathcal{F}_g$ into 3 semantical segments $\mathcal{F}_g^{c}$, $\mathcal{F}_g^{p}$ and $\mathcal{F}_g^{n}$;
            \State Compute prototypes of the 3 segments according to Eq. (\ref{proto_disentangle});
            \State Compute distance among the 3 segments (in Eq. (\ref{l_dec}));  \Comment{to get $\mathcal{L}_{dec}$}
            \If {$g < G-1$}
                \State Compute distance between ${f}_{b,g}^{c}$ and ${f}_{b,g+1}^{c}$ at every pair of adjacent granularities (in Eq. (\ref{ssdgc}));  \Comment{to get $S_{cs}$}
                \For {each parent class $q$ at granularity $g+1$}
                    \For {each sub-class $k$ at granularity $g$ that belongs to current parent class $q$}
                        \State Find all samples that belong to current sub-class $k$ and parent class $q$;
                        \State Compute sub-centroid $P_{k,q,g}^c$ of the common segments of these samples according to Eq. (\ref{proto_dsdgc});
                    \EndFor
                    
                    \State Compute distance between the sub-centroids $P_{q,g}^c$ of current parent class $q$ (in Eq. (\ref{dsdgc})); \Comment{to get $S_{cd}$}
                \EndFor
            %
            \EndIf
            \For {each class $k$ at granularity $g$}
                \State Find all samples that belong to class $k$;
                \State Compute the centroid $P_{k,g}^p$ of the specific segments of these samples according to Eq. (\ref{proto_dssgp});
            \EndFor
            \State Compute distance between the centroids $P_{g}^p$ (in Eq. (\ref{dssgp})); \Comment{to get $S_{p}$}
        \EndFor
        \State Aggregate and average all the distances to derive $\mathcal{L}_{dec}$, $S_{cs}$, $S_{cd}$ and $S_{p}$, respectively;
        \State Optimize $\mathcal{L}_{lf}$ and $\mathcal{L}_{c}$ in Eq. (\ref{llh}) and Eq. (\ref{l_c});
        \State Optimize $\mathcal{L}_{FS}$ in Eq. (\ref{lfs}) by minimizing $\mathcal{L}_{dec}$ and $S_{p}$, and maximizing $S_{cs}$ and $S_{cd}$;
    \EndFor
\end{algorithmic}
\end{algorithm*}
\vspace{-0.1in}

The pseudocode for the proposed FSDG approach is presented in Algorithm \ref{alg_1}, where the core steps and formulas are elaborated in detail. 
It is worth noting that the aforementioned processes are parallelized to enhance GPU efficiency by applying the constraints on the covariance matrix.

\section{Experiments}
\label{4experiments}

We train and evaluate the FSDG on three FGDG datasets for image recognition, comparing its generalization performance with several state-of-the-art DG methods. Additionally, functional analyses substantiate the effectiveness of FSDG. Codes will be available at https://github.com/YuWLong666/FSDG.

\subsection{Datasets}\label{dataset}
\textbf{CUB-Paintings} consists of two domains, i.e., CUB-200-2011(\textbf{C}) \cite{cub} and CUB-200-Paintings(\textbf{P}) \cite{PAN}, with significant domain shifts. Both include four levels of granularity, encompassing 14 orders, 38 families, 122 genera, and 200 species. The former \textbf{C} has 11,788 images of real-world bird species, while \textbf{P} is a collection of 3,047 images, including watercolors, oil paintings, pencil drawings, stamps, and cartoons.

\textbf{CompCars} \cite{comcars} collected car images from two domains: Web (\textbf{W}) and Surveillance (\textbf{S}). Two levels of hierarchy, comprising 68 coarse classes and 281 fine classes, were constructed based on the models and makes of the cars.
A DL model tested on \textbf{S} may exhibit suboptimal performance since the images were sourced exclusively from a single frontal perspective and were significantly affected by varying weather and lighting conditions.

\textbf{Birds-31} incorporates three domains: CUB-200-2011 (\textbf{C}), NABirds (\textbf{N}) \cite{nabirds}, and iNaturalist2017 (\textbf{I}) \cite{inalist}. Literature \cite{PAN} performed a union operation on the categories from the three datasets and selected 31 fine-grained categories, resulting in image counts of 1,848, 2,988, and 2,857, respectively. Subsequently, using the same granularity division method as \textbf{I}, four granularity levels were established, namely, 4 orders, 16 families, 25 genera, and 31 species.

\begin{table}[t]   
  \centering
\caption{Classification accuracy (\%) on the Cub-Paintings (RN-50) with best results of DG methods highlighted in \textbf{bold}.}
  \label{cub-quan}%
  \setlength{\tabcolsep}{3.9mm}{
    \begin{tabular}{lcccc}
    \toprule
    Method & C→P   & P→C   & Avg   & Params \\
    \midrule
        PAN (DA) & 67.40  & 50.92  & 59.16  & 103M \\
    \midrule
    ERM \cite{erm}   & 54.94  & 35.67  & 45.31  & 24M \\
    ARM \cite{arm}    & 47.98  & 31.53  & 39.76  & 24M \\
    DANN \cite{dann}  & 54.05  & 37.09  & 45.57  & 24M \\
    MLDG \cite{mldg}   & 55.40  & 34.15  & 44.78  & 23M \\
    GroupDRO \cite{groupdro} & 54.94  & 35.67  & 45.31  & 23M \\
    CORAL \cite{coral} & 54.70  & 35.29  & 45.00  & 23M \\
    SagNet \cite{sagnet} & 56.33  & 36.71  & 46.52  & 24M \\
    MixStyle \cite{mixstyle}  & 52.97  & 28.44  & 40.71  & 23M \\
    Mixup \cite{mixup}  & 54.58  & 34.66  & 44.62  & 23M \\
    RIDG \cite{ridg}  & 36.41  & 24.11  & 30.26  & 24M \\
    SAGM \cite{sagm}  & 57.83  & 37.16  & 47.50  & 23M \\
    MIRO \cite{miro}  & 56.29  & 41.28  & 48.79  & 47M \\
        \midrule
    S-FGDG & 61.21  & 41.90  & 51.56  & 26M \\
    S-FGDG (+$\mathcal{L}_{lf}$) & 61.98  & 41.99  & 51.98  & 26M \\
    S-FSDG & \textbf{63.42}  & 44.87  & 54.14  & 26M \\
    \midrule
    FGDG  & 56.76  & 46.53  & 51.65  & 49M \\
    FGDG (+$\mathcal{L}_{lf}$) & 60.18  & 47.13  & 53.66  & 49M \\
    FSDG  & 61.84 & \textbf{49.46} & \textbf{55.65} & 49M \\
    \bottomrule
        \end{tabular}%
      \vspace{-0.1in}
}
\end{table}%

\subsection{Implementation and Evaluation}\label{Implementation}
The feature extractor $E$ can be instantiated with ResNet (RN) series \cite{resnet}, Vision Transformer(ViT) series \cite{vit}, and ASMLP series \cite{asmlp} backbones, respectively. 
The learning rate is initially set to 0.003 and follows a dynamic strategy, where the coefficient of the learning rate varies from 1 to 0.1 during the training process. 
All layers except the backbones are trained from scratch, and their learning rate is 10 times that of the backbone. 
The model is optimized using mini-batch SGD with a momentum of 0.9. 
A single 3090 GPU with a batch size fixed to 32 is used for training and evaluation. 
A progressive grid search strategy is employed to identify the optimal weights for our proposed optimization objectives. The search space is configured as a set [0.005, 0.01, 0.05, 0.1, 0.5, 1].
Typically, the coefficients are configured as $\lambda_{cs} = 0.05$, $\lambda_{cd} = 0.5$, and $\lambda_{p} = 1$. $r^c$ and $r^p$ are set to 0.5 and 0.3, respectively. 
We employ fine-grained classification accuracy as the evaluation criterion. All models are trained three times and evaluated for ten trials on the target domain each time. 
Due to FSDG retaining only the finest branch during inference, the number of parameters of the dual backbone model is the same as that of the single backbone model, which is 24M. 

Features from the extractor $E$ are fed into a Granularity Transition layer $T$, which is comprised of a CNN layer followed by Batch Normalization and ReLU activation. 
The convolutional layer is configured with 2048 input channels, 256 output channels, and a kernel size of $1 \times 1$. All backbones are pre-trained on the ImageNet dataset.
We adopt the pre-trained ViT with a patch size set to 16 and an embedding dimension set to 384. 
The output of the final ViT block is reshaped into a feature tensor by a CNN layer with a kernel size of $3 \times 3$ and an output channel number of 2048.
As for ASMLP, pre-trained with a patch size set to 4, an embedding dimension set to 96, and a shifting size set to 5, we insert a CNN layer with a kernel size of $1 \times 1$ and an output channel number of 2048 between $E$ and $T$.

\begin{table}[t]  
\caption{Classification accuracy (\%) on the CompCars (RN-50) with best results of DG methods highlighted in \textbf{bold}.}
\label{cars-quan}%
  \centering
  \setlength{\tabcolsep}{3.9mm}{
    \begin{tabular}{lcccc}
    \toprule
    Method & W→S   & S→W   & Avg   & Params \\
    \midrule
    PAN (DA) & 47.05  & 15.57  & 31.31  & 103M \\
    \midrule
    ERM \cite{erm}  & 44.15  & 7.54  & 25.85  & 24M \\
    ARM \cite{arm}  & 20.25  & 4.74  & 12.50  & 24M \\
    DANN \cite{dann}  & 35.10  & 6.80  & 20.95  & 24M \\
    MLDG \cite{mldg}   & 44.94  & 7.56  & 26.25  & 23M \\
    GroupDRO \cite{groupdro} & 43.60  & 7.75  & 25.68  & 23M \\
    CORAL \cite{coral} & 43.05  & 7.97  & 25.51  & 23M \\
    SagNet \cite{sagnet} & 45.33  & 8.89  & 27.11  & 24M \\
    MixStyle \cite{mixstyle}  & 38.37  & 6.28  & 22.33  & 23M \\
    Mixup \cite{mixup}  & 43.07  & 7.56  & 25.32  & 23M \\
    RIDG \cite{ridg}  & 36.57  & 8.11  & 22.34  & 24M \\
    SAGM \cite{sagm}  & 49.55  & 8.58  & 29.07  & 23M \\
    MIRO \cite{miro}  & 46.01  & 7.88  & 26.95  & 47M \\
    \midrule
    S-FGDG & 50.87  & 8.26  & 29.56  & 26M \\
    S-FGDG (+$\mathcal{L}_{lf}$) & 52.09  & 9.58  & 30.83  & 26M \\
    S-FSDG & \textbf{53.44}  & 10.83  & \textbf{32.14}  & 26M \\
    \midrule
    FGDG  & 44.23  & 9.02  & 26.63  & 49M \\
    FGDG (+$\mathcal{L}_{lf}$) & 49.69  & 11.08  & 30.39  & 49M \\
    FSDG  & 51.78 & \textbf{11.30} & 31.54 & 49M \\
    \bottomrule
        \end{tabular}%
\vspace{-0.1in}
}
\end{table}%

\subsection{Main Results}\label{Results}

We compare our method with various DG methods, 
including ERM \cite{erm}, ARM \cite{arm} , DANN \cite{dann}, MLDG \cite{mldg}, GroupDRO \cite{groupdro}, CORAL \cite{coral}, SagNet \cite{sagnet}, MixStyle \cite{mixstyle} , Mixup \cite{mixup}, RIDG \cite{ridg}, SAGM \cite{sagm}, and MIRO \cite{miro}, based on their official codes and the DomainBed platform \cite{domainbed}. 
The $\mathcal{L}_{lf}$ in this context is referenced from PAN \cite{PAN}, although it was originally employed in a domain adversarial adaptation scenario rather than for FS in FSDG problems here. 
We separately present the performance of the baseline FGDG model, FGDG (+$\mathcal{L}_{lf}$) (i.e., replace $\mathcal{L}_{f}$ in Eq. (\ref{zonglossfgdg}) with $\mathcal{L}_{lf}$), and FSDG.
We also deploy the FSDG in a single backbone mode (prefixed with S-) to compare it with single-backbone DG methods.
The results are shown in Tables ~\ref{cub-quan}, \ref{cars-quan}, and \ref{birds31-quan}.

\textbf{On CUB-Paintings}, as in Table \ref{cub-quan}, our method performs best across DG competitors. 
With equivalent training parameter scales, both dual and single-backbone FSDGs outperform their respective second-best competitors (i.e., MIRO and SAGM) by 6.64\% and 6.86\%. 
In single and dual backbone scenarios, FSDGs exhibit substantial improvements of 2.16\% and 1.99\% over FGDG (+$\mathcal{L}_{lf}$) models, and 2.58\% and 4\% over the baselines, respectively. 
These results validate the superiority of FSDG, which enhances the assistance of coarse to fine granularity and identifies more generalizable fine-grained features. 
Common features can, to some extent, represent the invariance of species. 
They serve as intrinsic elements for classification and are beneficial to generalization. 
Other DG methods exhibit subpar generalization performance, revealing their vulnerability to fine-grained features.

\begin{table*}[!th]   
      \caption{Classification accuracy (\%) on the Birds-31 (RN-50) with best results of DG methods highlighted in \textbf{bold}.}
  \label{birds31-quan}%
  \centering
  \setlength{\tabcolsep}{5mm}{
    \begin{tabular}{lcccccccc}
    \toprule
    Method & C→I   & C→N   & I→C   & I→N   & N→C   & N→I   & Avg   & Params \\
    \midrule
        PAN (DA) \cite{PAN} & 69.79  & 84.19  & 90.46  & 88.10  & 92.51  & 75.03  & 83.34  & 103M \\
    \midrule
    ERM \cite{erm}   & 54.64  & 72.93  & 85.01  & 74.97  & 86.10  & 62.51  & 72.69  & 24M \\
    ARM \cite{arm}   & 50.51  & 71.25  & 77.38  & 74.20  & 84.74  & 59.82  & 69.65  & 24M \\
    DANN \cite{dann}  & 52.75  & 71.82  & 80.79  & 73.59  & 85.55  & 61.53  & 71.01  & 24M \\
    MLDG \cite{mldg}  & 53.55  & 72.19  & 80.74  & 74.83  & 85.61  & 61.95  & 71.48  & 23M \\
    GroupDRO \cite{groupdro} & 52.61  & 70.78  & 81.87  & 74.40  & 86.26  & 61.32  & 71.21  & 23M \\
    CORAL \cite{coral} & 54.64  & 72.93  & 81.01  & 74.97  & 86.10  & 62.51  & 72.03  & 23M \\
    SagNet \cite{sagnet} & 53.66  & 71.75  & 81.39  & 74.13  & 85.66  & 62.06  & 71.44  & 24M \\
    MixStyle \cite{mixstyle} & 49.95  & 69.04  & 74.46  & 68.34  & 83.60  & 57.12  & 67.09  & 23M \\
    Mixup \cite{mixup} & 52.36  & 71.65  & 82.36  & 75.17  & 85.61  & 62.34  & 71.58  & 23M \\
    RIDG \cite{ridg}  & 47.15  & 66.71  & 82.47  & 73.63  & 85.77  & 60.98  & 69.45  & 24M \\
    SAGM \cite{sagm}  & 54.04  & 73.63  & 82.96  & 77.01  & 87.88  & 63.49  & 73.17  & 23M \\
    MIRO \cite{miro}  & 54.39  & 74.87  & 82.36  & 75.34  & 86.42  & 62.48  & 72.64  & 47M \\
    \midrule
    S-FGDG & 65.24  & 81.73  & 88.66  & 84.90  & 90.71  & 72.30  & 80.59  & 26M \\
    S-FGDG (+$\mathcal{L}_{lf}$) & 65.18  & 81.46  & 88.57  & 84.91  & 90.53  & 72.05  & 80.45  & 26M \\
    S-FSDG & 63.66  & 82.43  & 89.56  & 85.80  & 92.03  & 72.91  & 81.06  & 26M \\
    \midrule
    FGDG  & 64.20  & 81.47  & 88.13  & 84.17  & 91.16  & 71.15  & 80.05  & 49M \\
    FGDG (+$\mathcal{L}_{lf}$) & 66.43  & 83.07  & 89.84  & 86.62  & 91.86  & 73.88  & 81.95  & 49M \\
    FSDG  & \textbf{66.32} & \textbf{83.71} & \textbf{90.69} & \textbf{87.36} & \textbf{91.95} & \textbf{74.20} & \textbf{82.37} & 49M \\
    \bottomrule
        \end{tabular}%
          \vspace{-0.05in}
}
\end{table*}%

\begin{table}[t]
\centering
\caption{Classification accuracy (\%) of various backbones with different depths on the Cub-Paintings dataset. RN and MLP represent ResNet and ASMLP backbones, respectively. T and S are abbreviations for Tiny and Small, respectively.}
\label{multibkb}%
\begin{small}
  \centering
    \begin{tabular}{llcccc}
    \toprule
    Backbone & Method & C→P   & P→C   & Avg   & Params \\
    \midrule
    \multirow{3}[2]{*}{RN-101} & FGDG  & 58.73    & 46.91   &52.82  & 87M \\
          & FGDG (+$\mathcal{L}_{lf}$) & 62.46  & 49.26  & 55.86  & 87M \\
          & FSDG  & 64.64  & 49.86  & 57.25  & 87M \\
    \midrule
    \multirow{3}[2]{*}{ViT-T} & FGDG  & 58.80  & 44.61  & 51.71  & 20M \\
          & FGDG (+$\mathcal{L}_{lf}$) & 60.37  & 48.69  & 54.23  & 20M \\
          & FSDG  & 60.71  & 48.82  & 54.76  & 20M \\
    \midrule
    \multirow{3}[2]{*}{ViT-S} & FGDG  & 67.93  & 64.86  & 66.40  & 60M \\
          & FGDG (+$\mathcal{L}_{lf}$) & 69.44  & 66.26  & 67.85  & 60M \\
          & FSDG  & 70.26  & 66.25  & 68.26  & 60M \\
    \midrule
    \multirow{3}[2]{*}{MLP-T} & FGDG  & 56.66  & 46.41  & 51.54  & 60M \\
          & FGDG (+$\mathcal{L}_{lf}$) & 58.38  & 47.96  & 53.17  & 60M \\
          & FSDG  & 60.90  & 50.51  & 55.70  & 60M \\
    \midrule
    \multirow{3}[2]{*}{MLP-S} & FGDG  & 58.85  & 49.42  & 54.14  & 103M \\
          & FGDG (+$\mathcal{L}_{lf}$) & 60.65  & 51.10  & 55.87  & 103M \\
          & FSDG  & 63.67   & 53.84  &  58.75     & 103M \\

\bottomrule    
\end{tabular}%
\vspace{-0.1in}
\end{small}
\end{table}%

\textbf{On ComCars}, as in Table \ref{cars-quan}, we obtain results with similar trends. FSDGs outperform second-best competitors by 3.07\% and 4.59\%, surpass baselines by 2.58\% and 4.91\%, and also outperform FGDG (+$\mathcal{L}_{lf}$) by 1.31\% and 1.15\%, respectively. 
The best results even surpass the PAN model. 
Single-backbone models achieve higher performance than dual-backbone models, which can be attributed to the substantial differences between the two domains. 
It is a challenge to generalize multiple levels of clean knowledge learned by dual backbones to a low-quality single-angle domain \textbf{S} when the model is trained on the high-quality multi-angle domain \textbf{W}.

\textbf{On Birds-31}, as in Table \ref{birds31-quan}, our method achieves the highest accuracy, surpassing the second-best method, SAGM, by approximately 9.2\%. Under the dual backbone configuration, FSDG outperforms the baseline by 2.32\%. 
On average, across three datasets, FSDG outperforms the second-best approach by up to 6.2\%.

\begin{table}[t]
\centering
\caption{Ablation study on the losses (i.e., $\mathcal{L}_{lf}$, $\mathcal{L}_{cs}$, $\mathcal{L}_{p}$, $\mathcal{L}_{cd}$, and $\mathcal{L}_{dec}$). The classification accuracy (\%) is presented.}
\label{ablation}
  \setlength{\tabcolsep}{2.7mm}{
    \begin{tabular}{ccccc|ccc}
    \toprule
     $\mathcal{L}_{lf}$   & $\mathcal{L}_{cs}$    & $\mathcal{L}_{p}$    & $\mathcal{L}_{cd}$ & $\mathcal{L}_{dec}$   & C→P   & P→C   & Avg \\
    \midrule
                    &       &       &     &             & 56.76  & 46.53  & 51.65  \\
     \checkmark     &       &       &      &            & 60.18  & 47.13  & 53.66  \\
     \checkmark     & \checkmark  &       &  &          & 61.26  & 47.76  & 54.51  \\
     \checkmark     &    & \checkmark     &   &         & 60.98  & 48.07  & 54.52  \\
     \checkmark  &    &    & \checkmark      &          & 60.85  & 47.52  & 54.18  \\
     \checkmark  &    &    &       &  \checkmark        & 60.68  & 47.51  & 54.09  \\
     \checkmark  & \checkmark  & \checkmark  &  &       & 61.18  & 48.55  & 54.87  \\
     \checkmark  & \checkmark  &  & \checkmark   &      & 61.66  & 47.77  & 54.71  \\
     \checkmark  &   & \checkmark  & \checkmark  &      & 60.60  & 48.50  & 54.55  \\
     & \checkmark  & \checkmark    & \checkmark  &      & 58.96  & 47.14  & 53.05  \\
     \checkmark &\checkmark &\checkmark &\checkmark   & & 61.77   & 48.43  &55.10  \\
     \checkmark &\checkmark &\checkmark &\checkmark &\checkmark &\textbf{61.84}  & \textbf{49.46}  &\textbf{55.65}  \\
    \bottomrule
    \end{tabular}
\vspace{-0.05in}
}
\end{table}

\textbf{Experiments on various backbones and depths.}
We conduct experiments on various backbones, including RN-50, RN-101, ViT-Tiny, ViT-Small, ASMLP-Tiny, and ASMLP-Small, as shown in Table \ref{multibkb}. 
Our method exhibits varying degrees of improvement across different architectures, ranging from 1.9 to 4.6\%. 
Upon replacing the backbone with ViT-S, FSDG achieves a performance of 68.26\%, surpassing PAN (DA) by 9.1\%, showcasing the superiority of Transformer architecture.
FS method achieves the highest performance improvement on ASMLP (4.6\%).
These results validate the robustness of the FSDG method across various backbone architectures.

\subsection{Analysis}\label{Analysis}

\textbf{Ablation study on the losses.} 
We present the outcomes of models optimized under various combinations of losses.
As shown in Table \ref{ablation}, each loss and their combinations contribute to performance improvement compared to the baseline FGDG and FGDG (+$\mathcal{L}_{lf}$) model (the first and second rows). 
Rows 3, 4, and 5 respectively demonstrate the increase in performance when each of the constraints is individually incorporated. 
Compared to row 2, which lacks feature partitioning and $\mathcal{L}_{dec}$, their inclusion results in generalization improvements of 0.85\%, 0.52\%, and 0.86\%. 
When all three constraints act in concert (second-to-last row), the generalization is enhanced by 1.44\%. 
The contrast between row 6 and row 2 reveals that the sole introduction of $\mathcal{L}_{dec}$ yields a 0.43\% improvement in generalization. 
When $\mathcal{L}_{dec}$ is optimized in synergy with the commonality and specificity constraints (last row), the model’s generalization is elevated to 55.65\%, marking a 1.99\% improvement. 
The addition of $\mathcal{L}_{dec}$ to the combined commonality and specificity constraints (second-to-last row) further boosts generalization by an additional 0.55\%.
These results all demonstrate the importance and effectiveness of the FS method proposed in this paper.

\begin{figure*}[!th]
\begin{center}
    {\includegraphics[width=1.5\columnwidth]{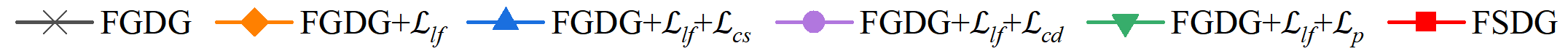}
    }\\
    \vspace{-0.1in}
    \subfloat[Similarity Value $S_{cs}$]{
        \includegraphics[width=0.587\columnwidth]{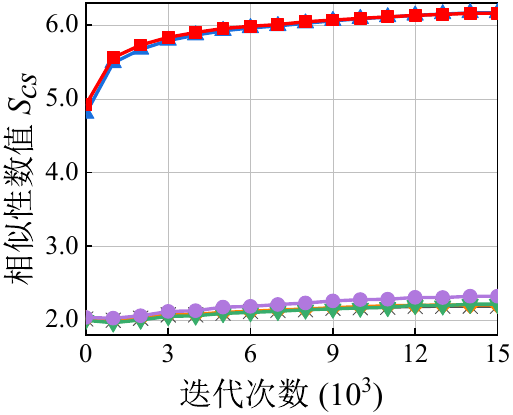}
        \label{scs}
        }
        \hspace{0.2in}
    \subfloat[Similarity Value $S_{cd}$]{
        \includegraphics[width=0.602\columnwidth]{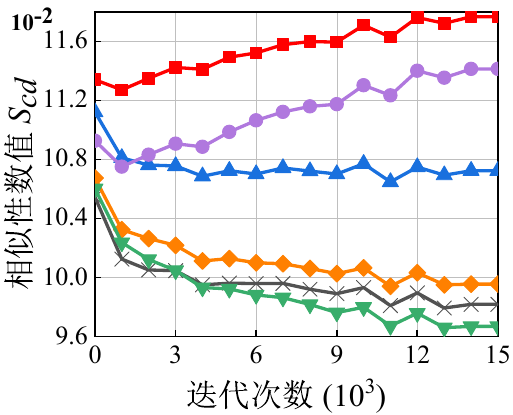}
        \label{scd}
        }
        \hspace{0.2in}
    \subfloat[Similarity Value $S_{p}$]{
        \includegraphics[width=0.595\columnwidth]{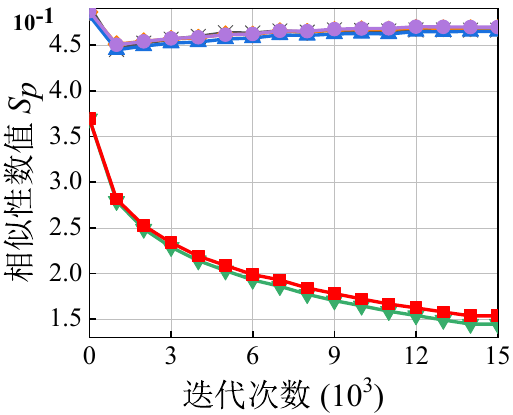}
        \label{sp}
        }
    \caption{Distance analyses of commonality and specificity. Under various combinations of losses, the distances among common features, (a) $S_{cs}$ and (b) $S_{cd}$, and the similarity of the specific parts, (c) $S_{p}$, are computed to illustrate the effectiveness of FS.}
    \label{similarityfig}
\end{center}
\vspace{-0.15in}
\end{figure*}

\textbf{Commonality alignment.} 
Fig. \ref{scs} and \ref{scd} illustrate how the similarities of common features, as depicted in Eq. (\ref{ssdgc}) and (\ref{dsdgc}), evolve throughout the training process under different loss functions. The similarity $S_{cs}$ significantly increases with the introduction of $\mathcal{L}_{cs}$. $\mathcal{L}_{cs}$ forces common parts at different granularities to converge. 
In other words, features of the common portion can extract the same information for a given sample, ensuring alignment of the shared concepts across various granularities.

$S_{cd}$ signifies the distance among sub-centroids' commonalities within a particular parent category. Fig. \ref{scd} shows that $\mathcal{L}_{cd}$ compels an increase in the similarity among different sub-categories. 
It can be observed that models lacking alignment tend to perceive common concepts as dissimilar. 
The network fails to capture the commonalities among sub-classes.
The reason behind this is that the network without alignment primarily fits discriminative features at a single granularity level, neglecting the shared characteristics among categories. 
Undoubtedly, this leads to the introduction of spurious discriminative features and a reduction in generalization ability.
Additionally, we observe a similar effect with $\mathcal{L}_{cs}$. The inclusion of $\mathcal{L}_{cs}$ also increases $S_{cd}$. 
When both losses are included, $S_{cd}$ reaches its optimal value, highlighting the necessity of employing dual common feature alignment constraints. Overall, to a certain extent, $\mathcal{L}_{cs}$ and $\mathcal{L}_{cd}$ have clustered the sub-categories and learned their common features from the perspective of FS.

\begin{figure}[t]
\begin{center}
    \centerline{\includegraphics[width=0.75\columnwidth]{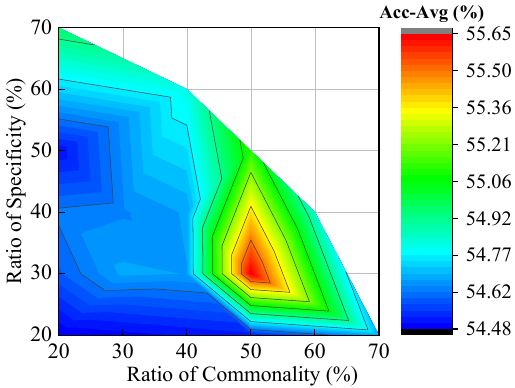}}
    \caption{Analyses of the proportions of common, specific, and confounding components. 
}
    \label{ratio}
\end{center}
\vspace{-0.2in}
\end{figure}

\textbf{Specificity alignment.} 
$\mathcal{L}_{p}$ forces the specific segment of different categories to be as far apart as possible. As depicted in Fig. \ref{sp}, $\mathcal{L}_{p}$ significantly diminishes the similarity among individuals from different categories, and this gap continues to widen further as the training advances, thereby affirming the efficacy of the devised approach.
Moreover, the diminution in specific feature similarity is conducive to amplifying discriminability among category representations, thereby fortifying the capability for fine-grained recognition. 

\textbf{Ratio impact.} 
FSDG reaches its peak when the proportions of commonality, specificity, and confounding are controlled to be 5:3:2, as Fig. \ref{ratio} shows. 
As the ratios of commonality and specificity decrease, the performance in FGDG drops to a minimum.
Compared to specificity, commonality has a greater impact on generalization.
The constraints on the common and specific parts contribute to the improved generalization. 
Even the lowest performance (54.48\%) at the point where common and specific components only account for 20\% each is still higher than the case without the $\mathcal{L}_{FS}$ constraint, which is 53.66\% in Table \ref{cub-quan}. 
This is because, in this study, we impose no strict constraints on the confounding part, which means increasing the ratio of the confounding part tends to resemble the original FGDG model. 
Moreover, performance is degraded when $r^n$ is configured as 0, highlighting the necessity of isolating the confounding part from the commonality and specificity at the feature level.
These results all validate the necessity of structuring features into the three segments.

\begin{table}[t]
\centering
\caption{Classification accuracy (\%) and Gpu Hours of various distance metrics. The GPU Hours are tested on conditions: single 3090 GPU, batch size 32, and epoch 1.}
\label{hsic}%
    \setlength{\tabcolsep}{4.4mm}{
    \begin{tabular}{ccccc}
    \toprule
    Method & C→P   & P→C   & Avg   & Gpu Hours \\
    \midrule
    HSIC        & 61.10  & 47.97  & 54.54  & 0.34  \\
    Euclidean   & 60.93  & 47.90  & 54.41  & 0.05  \\
    Cosine      &\textbf{61.84}  & \textbf{49.46}  &\textbf{55.65}  & 0.06  \\
    \bottomrule
\end{tabular}%
}
\end{table}%

\begin{table}[t]
\caption{Classification accuracy (\%) on various network architectures of the Granularity Transition Layers $T$. BN represents Batch Normalization. Additional configurations for CNN include a stride of 1 and a padding of 1.}
\label{tnet}%
\centering
\setlength{\tabcolsep}{3mm}{
\begin{tabular}{cccc}
\toprule
    Network Architecture & C→P   & P→C   & Avg \\
    \midrule
    \multicolumn{1}{c}{Conv 1$\times$1$\times$ $d$, BN, ReLU, Dropout,} & \multirow{4}{*}{59.10}  & \multirow{4}{*}{43.59}  & \multirow{4}{*}{51.34}  \\
    \multicolumn{1}{c}{Conv 1$\times$1$\times$ $d$, BN, ReLU, Dropout,} \\
    \multicolumn{1}{c}{Conv 1$\times$1$\times$256, BN, ReLU, } \\
    \multicolumn{1}{c}{ShortCut} \\
    \midrule
    
    \multicolumn{1}{c}{Conv 1$\times$1$\times$ $d$, BN, ReLU, Dropout,} & \multirow{3}{*}{60.48}  & \multirow{3}{*}{46.43}  & \multirow{3}{*}{53.45}  \\
    \multicolumn{1}{c}{Conv 1$\times$1$\times$256, BN, ReLU,} \\
    \multicolumn{1}{c}{ShortCut} \\
    \midrule
    \multicolumn{1}{c}{Conv 1$\times$1$\times$256, BN, ReLU,} & \multirow{2}{*}{61.50}  & \multirow{2}{*}{48.76}  & \multirow{2}{*}{55.13}  \\
    \multicolumn{1}{c}{ShortCut} \\
    \midrule
    
    \multicolumn{1}{c}{Conv 1$\times$1$\times$ $d$, BN, ReLU, Dropout,} & \multirow{3}{*}{59.05}  & \multirow{3}{*}{43.02}  & \multirow{3}{*}{51.03}  \\
    \multicolumn{1}{c}{Conv 1$\times$1 $\times$ $d$, BN, ReLU, Dropout,} \\
    \multicolumn{1}{c}{Conv 1$\times$ 1$\times$ 256, BN, ReLU} \\
    \midrule
    
    \multicolumn{1}{c}{Conv 1$\times$1$\times$$d$, BN, ReLU, Dropout,} & \multirow{2}{*}{60.43}  & \multirow{2}{*}{46.23}  & \multirow{2}{*}{53.33}  \\
    \multicolumn{1}{c}{Conv 1$\times$1$\times$256, BN, ReLU} \\

    \midrule
    \multicolumn{1}{c}{Conv 1$\times$1$\times$256, BN, ReLU} & \multirow{1}{*}{\textbf{61.84}}  & \multirow{1}{*}{\textbf{49.46}}  & \multirow{1}{*}{\textbf{55.65}}\\
\bottomrule
\end{tabular}%
}
\label{tab:addlabel}%
\vspace{-0.1in}
\end{table}%

\begin{table}[t] 
\centering
\caption{\textcolor{blue}{Classification accuracy (\%) of FGVC methods on FGDG task.}}
\label{tab.fgvc}%
\setlength{\tabcolsep}{6.9mm}{  
\begin{tabular}{lccc}
\toprule
    Method & C→P & P→C & Avg   \\
\midrule
    MPSA \cite{fgvc-mpsa}    & 59.07  & 36.31 & 47.69  \\
    SEF \cite{fgvc-sef}   & 58.09 &  35.63 &  46.86 \\
\bottomrule
\end{tabular}%
}
\end{table}%

\begin{table}[!t] 
\centering
\caption{\textcolor{blue}{Classification accuracy (\%) of in-domain and out-of-domain scenarios. 
}}
\label{tab.indomain}%
\setlength{\tabcolsep}{2.5mm}{  
\begin{tabular}{lcc}
\toprule
    Method & In-Domain Acc.   & Out-of-Domain Acc. (C→P)\\
\midrule
    S-FGDG (Baseline) & 87.71  & 61.21 \\
    S-FGDG (+$\mathcal{L}_{lf}$) & 86.98  & 61.98 \\
    S-FSDG (Ours) & 86.37  &  63.42 \\
\bottomrule
\end{tabular}%
\vspace{-0.1in}
}
\end{table}%

\textbf{Experiments on various distance measurements.} 
We conduct experiments with HSIC and Euclidean distance as alternatives of Cosine similarity, as shown in Table \ref{hsic}. 
From the perspectives of generalization and computational efficiency, we find that the performance achieved by Cosine exceeds that of other methods by approximately 1\%, while the computational time is slightly higher than the Euclidean by 0.01 hours, which is acceptable. 
Besides, Cosine represents the degree of orthogonality between two vectors, which meets our needs. 
Therefore, we opt to utilize Cosine similarity.

\textbf{Experiments on various architectures of the Granularity Transition Layer $T$.}
$T$ refines the globally shared features extracted by backbones, tailoring them specifically to each granularity branch. 
We configure various structures to construct $T$ with results presented in Table \ref{tnet}.
The simplest single-layer CNN achieves the best results. As the number of CNN layers increases, the model's performance declines, indicating a loss of useful information. 
Therefore, we choose the single-layer CNN as the Granularity Transition Layer.

\textbf{Comparison with FGVC methods.}
We conduct experiments to evaluate FGVC methods on the domain generalization task, with the results presented in Table \ref{tab.fgvc}. 
The generalization capability of FGVC methods is consistently lower than that of the proposed FSDG model. 
Compared to the generalization of the single-backbone S-FSDG (54.14\%), the generalization of the FGVC method is 6.87\% lower.
In pursuing higher fine-grained recognition accuracy, they inadvertently learn spurious correlations, which compromise their generalization performance. 
When sample distributions shift, such as changes in background or style, the recognition accuracy of these methods decreases significantly.

\begin{figure}[!th]
\begin{center}
\hspace{-0.1in}
\subfloat[Single optimization]{
    \includegraphics[width=0.489\columnwidth]{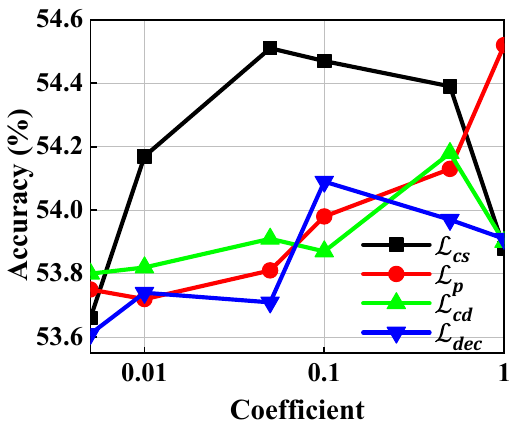}
    \label{coe-every}
    }
    \hspace{-0.168in}
\subfloat[Progressive optimization]{
    \includegraphics[width=0.492\columnwidth]{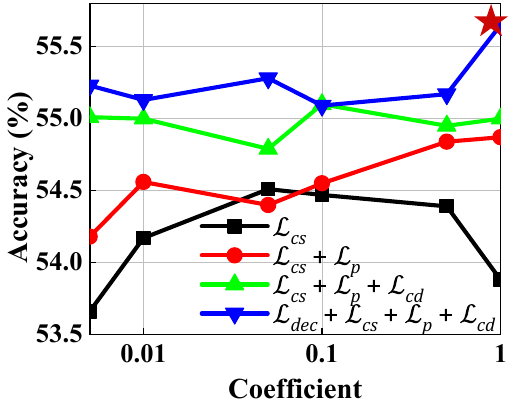}
    \label{coe-progress}
    }
\caption{Performance of various loss functions under different coefficients. (a) presents the performance of each loss. (b) is the progressive optimization of the 4 losses.}
\label{coe-losses}
\end{center}
\vspace{-0.1in}
\end{figure}

\textbf{Evaluation on in-domain and out-of-domain scenarios.}
We conduct the evaluation of the FSDG and baseline models on both in-domain and out-of-domain settings, as detailed in Table \ref{tab.indomain}. 
Three models are trained on the CUB-200-2011 dataset, with in-domain performance assessed on the validation subset and out-of-domain performance on the CUB-200-Painting domain. 
The results indicate that the FSDG model experiences a performance decrement of approximately 1.34\% in the in-domain setting while demonstrating a significant enhancement of 2.21\% in the out-of-domain setting. 
This outcome underscores that the enhancement in generalization brought about by FS is directly attributed to the model’s inherent generalization capabilities.
It effectively mitigates the influence of spurious correlations such as background and style, enabling the model to learn domain-invariant features.

\textbf{Analysis of coefficients on loss functions.}
We first conduct an optimal coefficient search for each loss. 
As shown in Fig. \ref{coe-every}, each loss possesses an optimal value within the search space and confers an enhancement in generalization.
Starting at $\lambda_{cs}$, we then progressively identify the optimal coefficient and fix it to search for the optimal point of the subsequent losses.
As depicted in Fig. \ref{coe-progress}, this strategy results in a steady improvement of the model’s performance, culminating in the optimal generalization performance (55.65\%).
Ultimately, the optimal set of parameters is obtained.

\begin{table*}[tbp]
    \centering
    \tabcaption{4-granularity labels of 8 category examples and the overlap of statistical top concepts between each category and the other 7 categories. The ratio of commonality shows the ratio between the Common and the All.}
    \label{tab:4glabel}
    \scalebox{1}{
    \setlength{\tabcolsep}{2.6mm}{
    \begin{tabular}{cccc|cccc|c|cccc|c}
    \toprule
    \multicolumn{4}{c|}{Category Examples} & FGDG  & FGDG  & FGDG  & FGDG  & FGDG & FSDG  & FSDG  & FSDG  & FSDG  & FSDG
    \\
        g=0  &   g=1    &   g=2    &  g=3     & All   & Com.   & Spe.   & Conf.   & Ratio Com.   & All   & Com.   & Spe.   & Conf.   & Ratio Com. \\
    \midrule
    8     & 5     & 3     & 3     & 94    & 36    & 35    & 23    & 38\%  & 61    & 35    & 13    & 13    & 57\% \\
    9     & 6     & 3     & 3     & 70    & 24    & 23    & 23    & 34\%  & 62    & 34    & 12    & 16    & 55\% \\
    10    & 5     & 3     & 3     & 94    & 39    & 32    & 23    & 41\%  & 70    & 55    & 8     & 7     & 79\% \\
    11    & 7     & 3     & 3     & 72    & 27    & 30    & 15    & 38\%  & 65    & 47    & 12    & 6     & 72\% \\
    12    & 8     & 3     & 3     & 85    & 29    & 32    & 24    & 34\%  & 58    & 41    & 7     & 10    & 71\% \\
    28    & 19    & 12    & 3     & 99    & 45    & 39    & 15    & 45\%  & 60    & 34    & 13    & 13    & 57\% \\
    29    & 19    & 12    & 3     & 96    & 37    & 33    & 26    & 39\%  & 64    & 44    & 9     & 11    & 69\% \\
    51    & 36    & 19    & 8     & 32    & 18    & 10    & 4     & 56\%  & 32    & 25    & 7     & 0     & 78\% \\
    \midrule
    \multicolumn{4}{c|}{Average}   & 80  & 32 & 29 & 19 & 40\%  & 59    & 40 & 10 & 9   & 68\% \\
    \bottomrule
    \end{tabular}
    }}
\vspace{-0.15in}
\end{table*}

\begin{figure*}[!th]
\begin{center}
    \subfloat[Ground Truth of Concept Overlap]{
        \includegraphics[width=0.53\columnwidth]{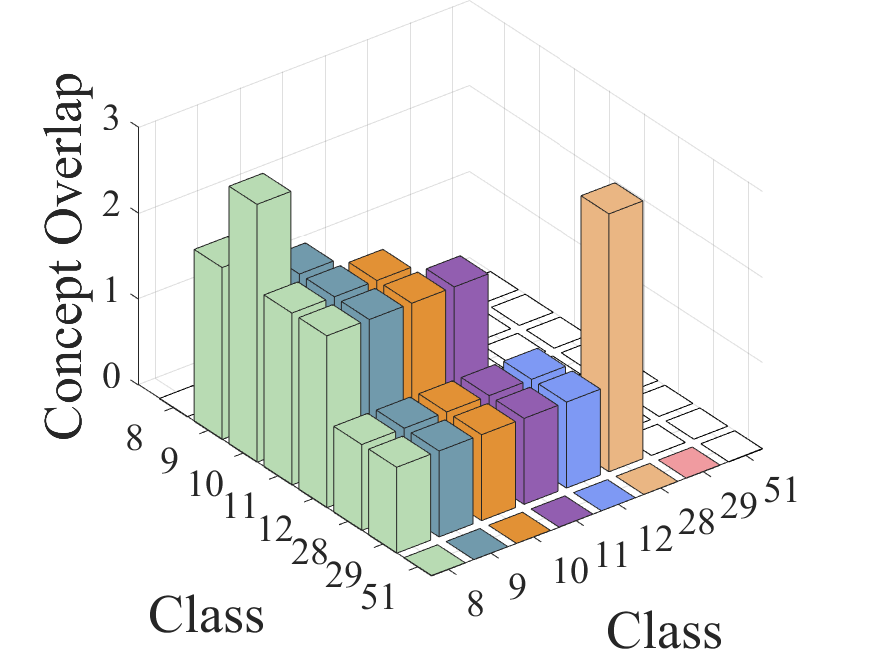}
        \label{fig:cogt}
        }
    \hspace{-7mm}
    \subfloat[Common]{
        \includegraphics[width=0.49\columnwidth]{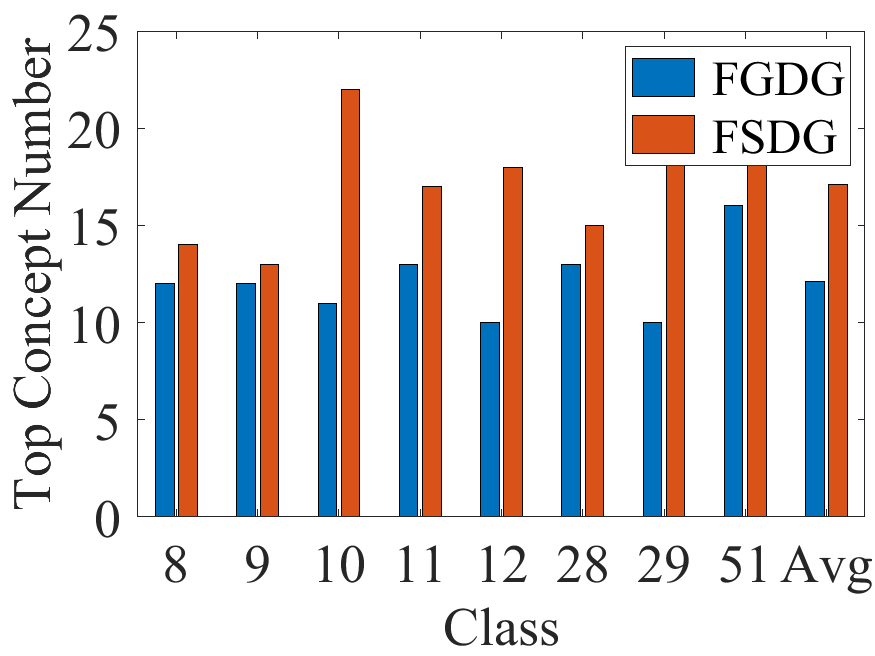}
        \label{fig:diagcom}
        }
    \subfloat[Specific]{
        \includegraphics[width=0.49\columnwidth]{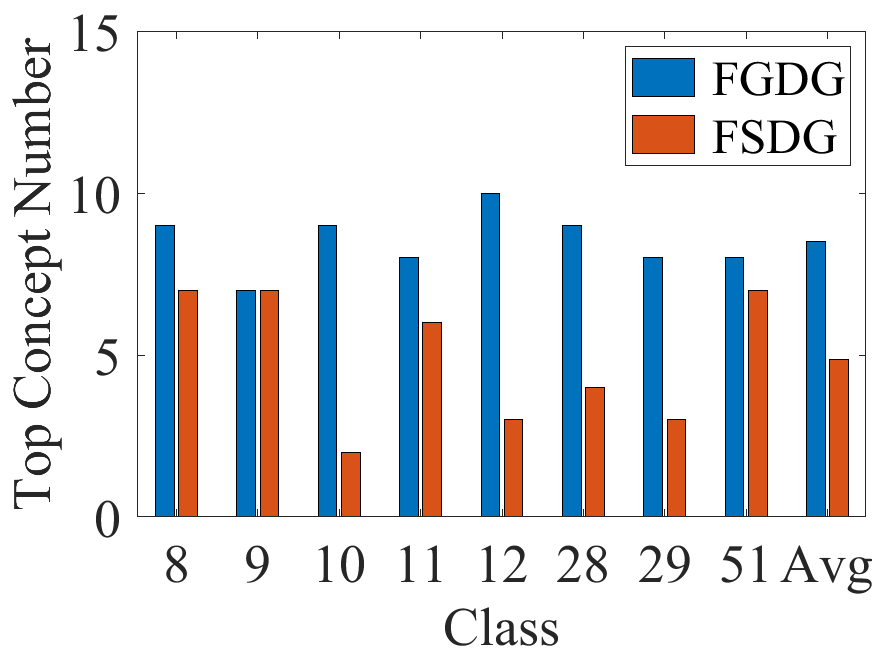}
        \label{fig:diagspe}
        }
    \subfloat[Confounding]{
        \includegraphics[width=0.49\columnwidth]{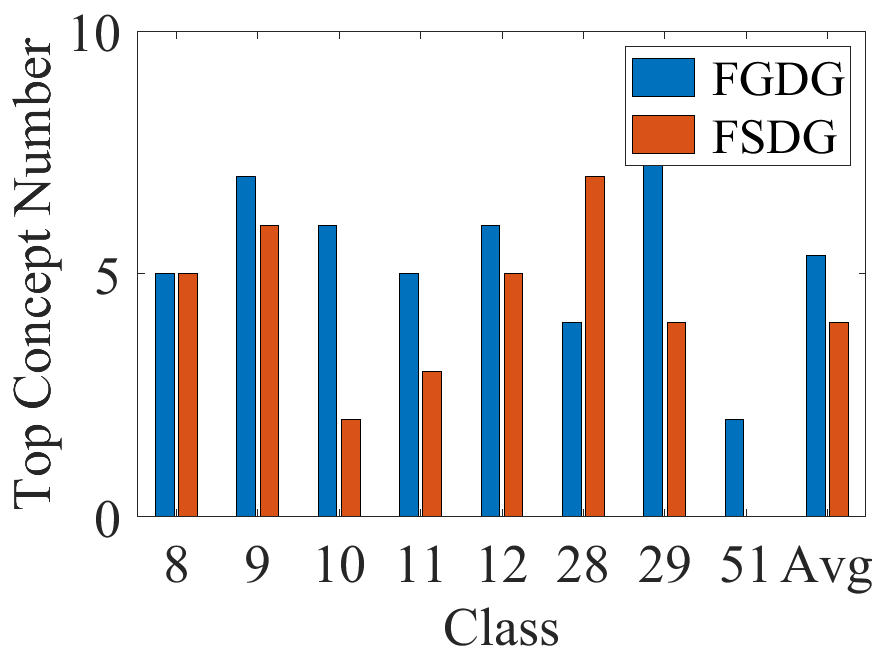}
        \label{fig:diagnoi}
        }\\   
    \caption{Confusion matrix of the ground truth of concept overlaps and the histograms of the most relevant concepts. (b), (c) and (d) plot the number of top concepts belonging to the three segments. Avg shows the average value among 8 categories.}
    \label{fig:diag}
\end{center}
\vspace{-0.2in}
\end{figure*}

\section{Explainability Analysis}\label{Sec-explain}

In this paper, we enhance the FGDG performance through the FS method with the integration of data and knowledge. 
The demonstrated enhancement primarily manifests in that the common part of the features, decoupled based on channel indices, primarily focuses on the commonalities and representing the invariance between categories, while the specific part focuses on learning discriminative characteristics.
In this section, we endeavor to validate that FS increases the explicit concept matching intensity between the shared concepts among categories and the indices of model channels, 
and the FSDG model embeds knowledge of a multi-granularity structure within its feature space.


We argue that categories that are closer together should activate more identical concepts in the common part, while classes that are farther apart activate fewer shared concepts in this segment. 
To demonstrate this argument, we randomly select a sequence of classes from the Cub-Paintings dataset, and their corresponding category labels at four granularities are shown in Table \ref{tab:4glabel}. 
We then define distances based on a multi-granular structure to roughly measure the similarity between fine-grained categories:
\begin{equation} \label{sclass}
    \mathcal{S}_{class}^{i, j}=d_i-\left\|\boldsymbol{c}_i-\boldsymbol{c}_j\right\|_0,
\end{equation}
where $\boldsymbol{c}_i$ is a class vector in which each element represents the class label of the according granularity (e.g., $\boldsymbol{c}_8=[8,5,3,3]^\top$), and $d_i$ reflects the dimension of $\boldsymbol{c}_i$, respectively. 

$\mathcal{S}_{class}$ is a discrete numerical value, with its maximum determined by the number of granularities in the hierarchical knowledge structure. 
$\mathcal{S}_{class}$ provides a ranking of the differences between different fine-grained categories. 
It should be noted that $\mathcal{S}_{class}=0$ does not imply that there is no commonality between two categories but rather indicates low similarity between them within the multi-granular structure. 
For example, in the Cub-Paintings dataset, all bird species share a consistent body shape, which is a form of commonality. 
However, this similarity might not be reflected within the current granularity structure.
Fig. \ref{fig:cogt} displays the confusion matrix of $\mathcal{S}_{class}$ between fine-grained category instances as listed in Table \ref{tab:4glabel}. Fig. \ref{fig:cogt} depicts the ground truth quantifying the difference ranks between categories.

\begin{figure*}[!th]
\begin{center}
    \subfloat[All of FGDG]{
        \includegraphics[width=0.538\columnwidth]{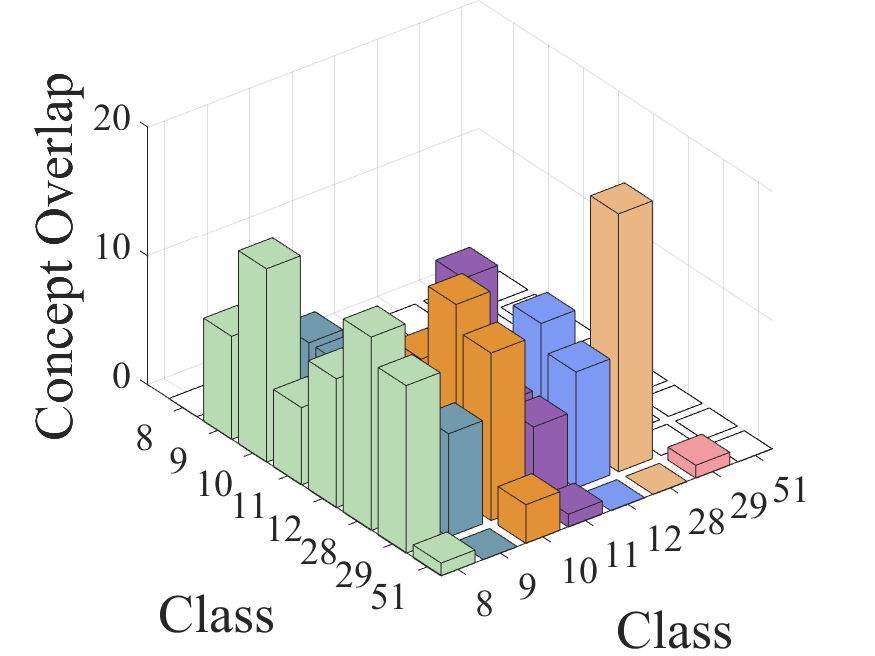}
        \label{allfgdg}
        }
    \hspace{-7.655mm}
    \subfloat[Common of FGDG]{
        \includegraphics[width=0.538\columnwidth]{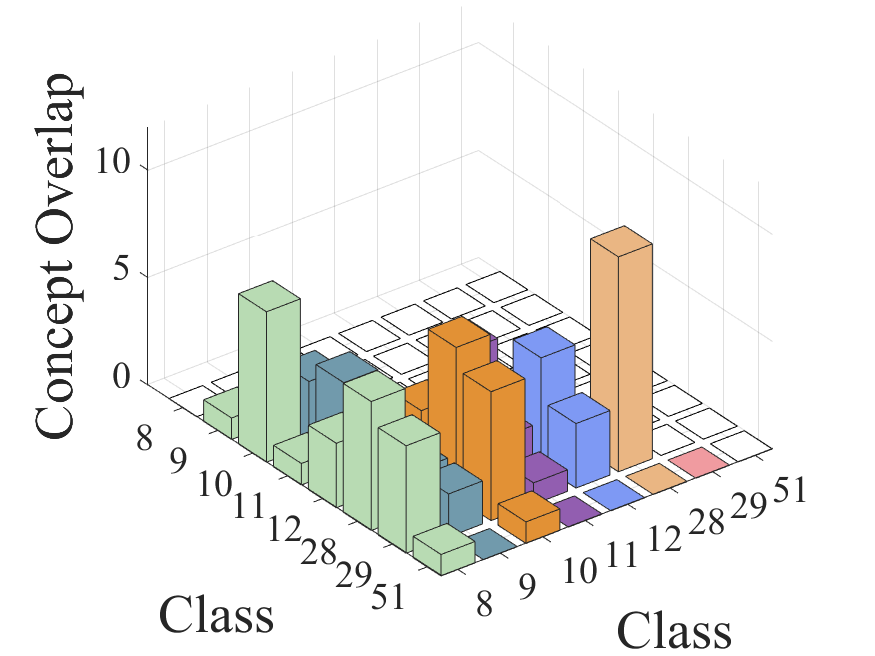}
        \label{comfgdg}
        }
    \hspace{-7.655mm}
    \subfloat[Specific of FGDG]{
        \includegraphics[width=0.538\columnwidth]{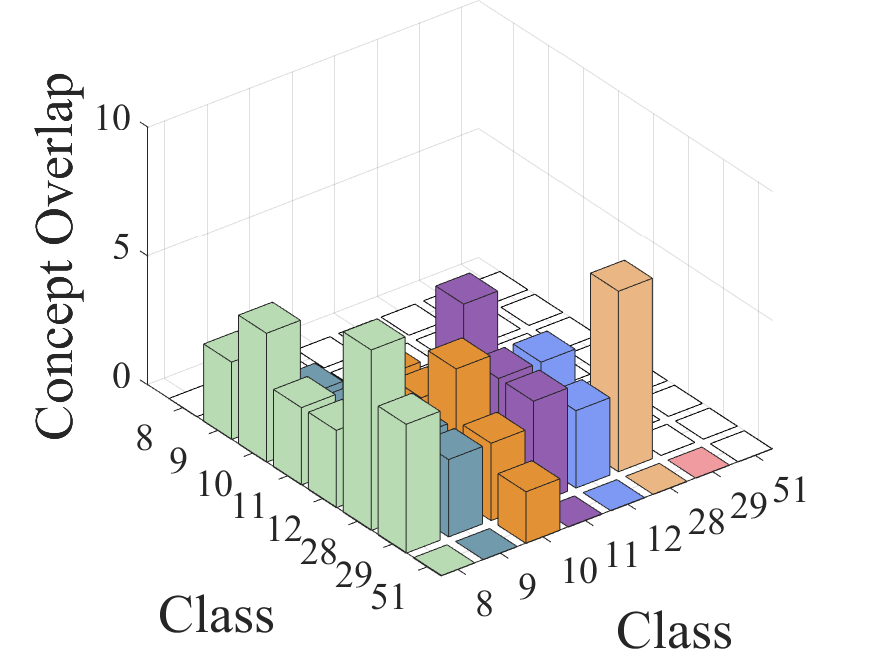}
        \label{spefgdg}
        }
    \hspace{-7.655mm}
    \subfloat[Confounding of FGDG]{
        \includegraphics[width=0.538\columnwidth]{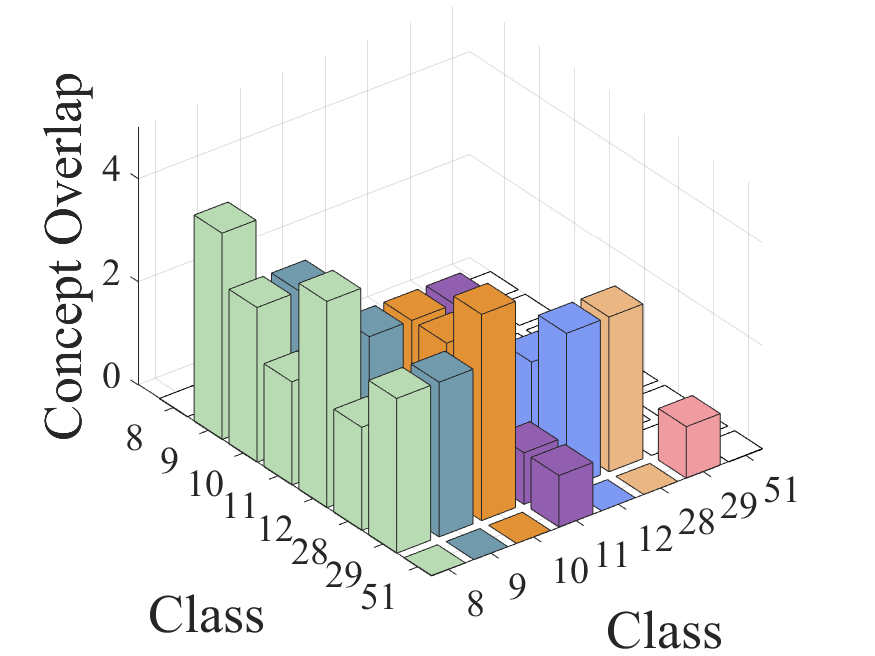}
        \label{noifgdg}
        }\\   
    \vspace{-2mm}
    \subfloat[All of FSDG]{
        \includegraphics[width=0.538\columnwidth]{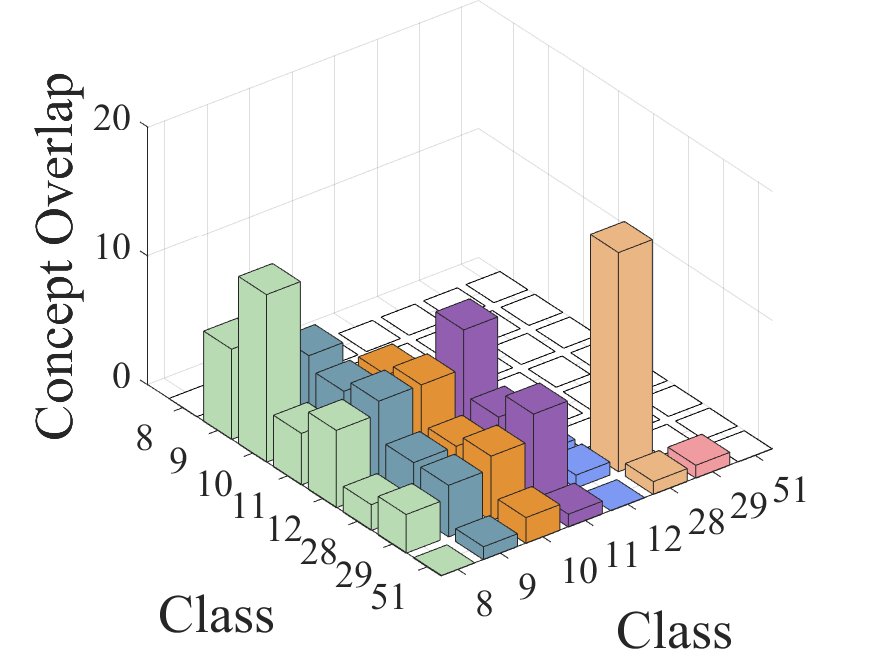}
        \label{allfsdg}
        }
    \hspace{-7.655mm}
    \subfloat[Common of FSDG]{
        \includegraphics[width=0.538\columnwidth]{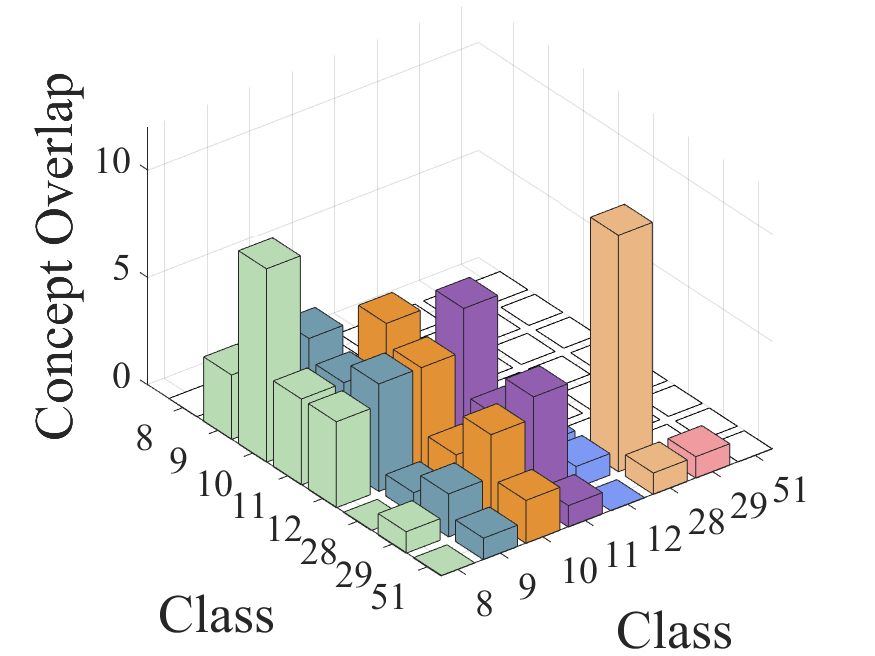}
        \label{comfsdg}
        }
    \hspace{-7.655mm}
    \subfloat[Specific of FSDG]{
        \includegraphics[width=0.538\columnwidth]{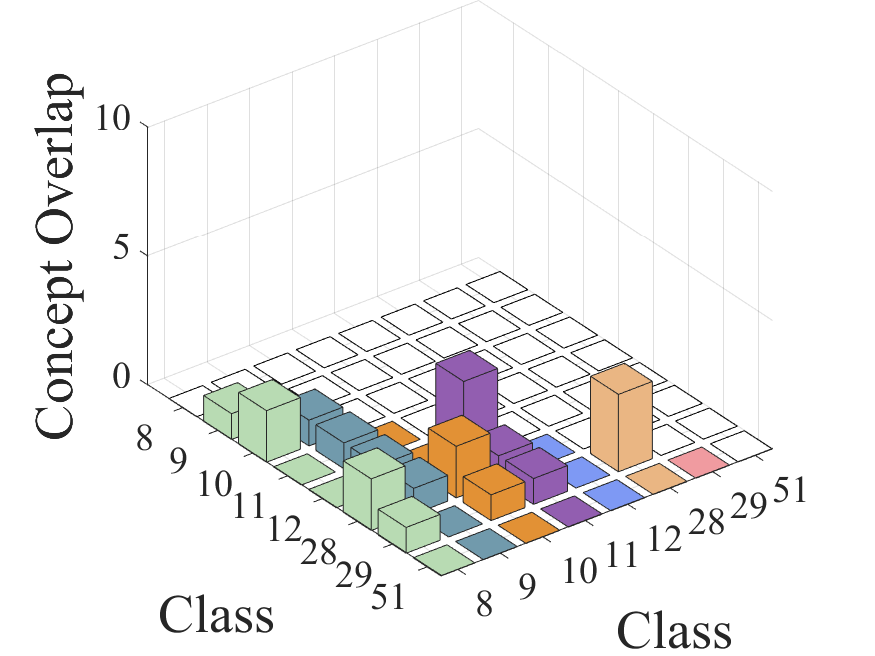}
        \label{spefsdg}
        }
    \hspace{-7.655mm}
    \subfloat[Confounding of FSDG]{
        \includegraphics[width=0.538\columnwidth]{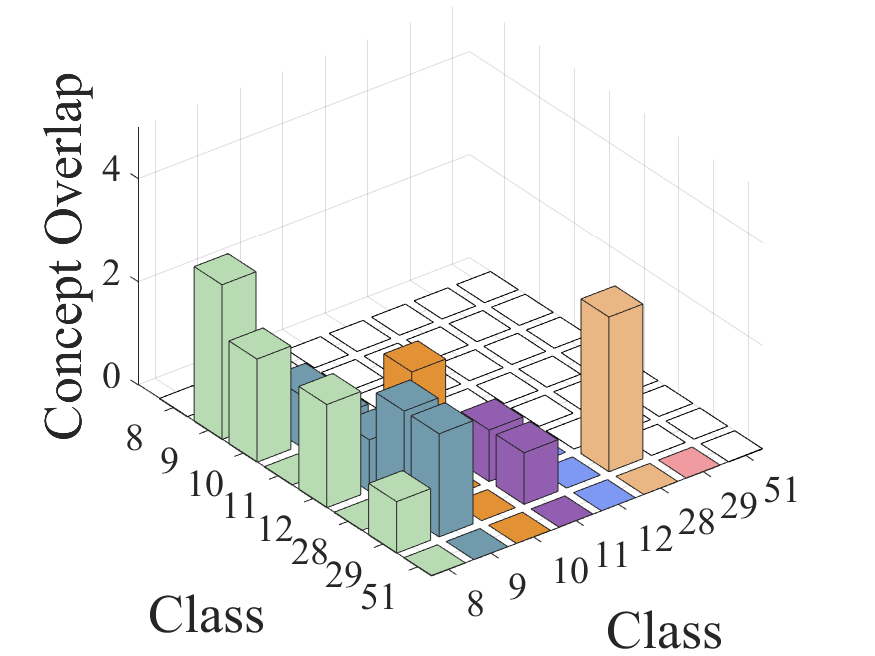}
        \label{noifsdg}
        }
    \caption{Confusion matrix of the overlap of relevant concepts between each pair of categories. The first row represents the concept overlaps of the baseline FGDG model, while the second row belongs to the proposed FSDG method. The first column denotes the overlap of all 26 top relevant concepts; the second column denotes those belonging to the common part; the third column denotes the specific part; and the fourth column denotes the confounding part. The computed confusion matrix is symmetric. To enhance visualization, we set the upper triangle of the confusion matrix to zero. The ground truth and the diagonal elements of the confusion matrices are separately plotted in Fig. \ref{fig:diag}.}
    \label{fig:alloverlap}
\end{center}
\vspace{-0.2in}
\end{figure*}

We then refine a Concept Relevance Propagation (CRP) technique \cite{naturecrp} to acquire the most relevant shared concepts activated by every two categories. 
Specifically, CRP first records the 40 most relevant samples and their relevance values with respect to each channel of the transition layer $T_{g=0}$. 
We further refine the CRP by statistically sorting all 256 channels for each class according to their relevance values. 
As each channel can represent a concept, this statistical outcome achieves relevance ranking for all 256 concepts concerning each class.
After that, the top 26 channels and their relevance values of each class are extracted to represent the most relevant concepts.
The reason why we choose 26 is that, for a single category, only approximately the top 10\% of channels are activated, while the remaining 90\% have activation levels less than one-tenth of the maximum activation level.

To quantitatively analyze the matching intensity between shared concepts and channel structure regions, we sum up the number of shared concepts between each category and the other seven categories, as shown in Table \ref{tab:4glabel}. 
By calculating the proportion of shared concepts within the common region relative to all shared concepts, 
it can be observed that the ratio of overlapping concepts that fall within the commonality segment for models without FS is 40\% (even lower than the 50\% ‘common’ ratio of the partitioning scheme based on a 5:3:2 ratio). 
Upon the incorporation of the FS method, this proportion increases to 68\%.
The proportion of these overlapping concepts within the specificity segment decreases (from 36\% to 17\%), indicating an increased focus on category-specific concepts in this segment.
Besides, Fig. \ref{fig:diagcom}-\ref{fig:diagnoi} illustrate the distributions of the top 26 relevant concepts in three parts. 
FS leads to an increase in the number of concepts activated in the common region while reducing the activation in the other two regions. 
These findings validate that FS successfully confines the semantical commonalities among categories to the first segment, while the second segment recognizes concepts unique to each category.
When predicting the given samples, the FSDG model pays more attention to the general semantics through its common segment.

\begin{table}[t]
\centering
\caption{
Spearman Rank Correlation Coefficient of concept overlap between regional features of various models and the ground truth. The second column represents the value between all features. The third column is the comparison between ground truth and the common segment.
}
\label{pearson}
  \setlength{\tabcolsep}{4.6mm}{
    \begin{tabular}{c |cccc}
    \toprule
    FGDG & All & Com. & Spe. & Conf.\\
    \midrule
    Ground Truth & 0.62 & 0.52 & 0.55 & 0.67 \\
    \midrule
    FSDG & All & Com. & Spe. & Conf. \\
    \midrule
    Ground Truth & 0.88 & 0.83 & 0.50 & 0.60 \\
    \bottomrule
    \end{tabular}
\vspace{-0.1in}
}
\end{table}

We further substantiate that the FSDG model embeds a multi-granularity structure into its feature embeddings, enabling it to perceive the similarities between categories.
Intuitively, we argue that closer categories should share more common concepts, meaning that there should be a higher overlap in the most relevant activated concepts. 
We compute the confusion matrix of the overlap of relevant concepts between each pair of category examples, as illustrated in Fig. \ref{fig:alloverlap}. 
Considering that the FGDG model does not undergo partitioned training, we compare the learning capabilities of the FGDG and FSDG models with respect to the multi-granularity structure across all features (i.e. Fig. \ref{allfgdg} and Fig. \ref{allfsdg}).
Compared to Fig. \ref{allfgdg}, the overall shape of the concept overlap activated by FS (Fig. \ref{allfsdg}) closely resembles the ground truth (Fig. \ref{fig:cogt}). 
We proceed to calculate the Spearman Rank Correlation Coefficient for the overlapping concepts between different segments and the ground truth for both models, as presented in Table \ref{pearson}. 
It is observed that the FSDG model achieves a correlation coefficient of 0.88 in learning explicit category similarity, whereas the FGDG model only reaches 0.62. 
The common segment of the FSDG model also exhibits a correlation coefficient of 0.83, enabling the learning of general features between categories. 
The correlation between the other two segments and the ground truth is relatively weak. 
The FGDG model, without FS, overlooks the relationships between categories, resulting in distant classes having a considerable overlap in activated shared concepts, which is inaccurate.
These results demonstrate that the FSDG model embeds a perception of multi-granularity structure within the feature space.
When identifying objects, it is capable of perceiving the distances between fine-grained categories.

When contrasting the distribution of shared concepts in the specific region, as shown in Fig. \ref{spefsdg}, we observe that FS elicits a sparser distribution of shared concepts.
The correlation between these features and the multi-granularity category similarity ground truth is only 0.5.
This indicates that this region focuses more on specificities, aligning with the initial assumptions of the model. 
This phenomenon also occurs in the confounding part. 
Besides, we do not impose constraints on the confounding part, a natural discussion arises as to whether applying constraints only to the common and specific parts would increase more activations in the confounding part. 
In contrast, by analyzing Fig. \ref{fig:diagnoi} and Fig. \ref{noifsdg}, we find that the number of top activations in the confounding part is decreased. 
All of these results demonstrate that FS embeds a multi-granularity structure into the feature space and achieves feature functionalization in accordance with the objectives. It directs the model to focus more on capturing general and specific features between categories, therefore improving generalization.
This also indicates that we, to some extent, pre-define the functionalities of the channels, enhancing the model's internal transparency and explainability.


\section{Conclusion}

In this paper, we assimilate insights from cognitive psychology to advance a feature structuralization framework to tackle the fine-grained domain generalization challenge.
FS explicitly embeds structured commonalities and specificities into DL models by integrating data and multi-granularity knowledge.
The disentanglement of learned features into common, specific, and confounding segments and their decorrelation, the feature alignment constraints facilitated by three optimization functions, and the prediction calibration term collaboratively contribute to the construction of the FSDG model.
Extensive experiments demonstrate a notable enhancement in FGDG performance. 
Furthermore, the explainability analysis confirms the validity of FS.

Despite FSDG's systemic efficacy, it brings forth new challenges and prompts intriguing questions.
For example, the pre-constructed granularity structure is used to achieve semantic feature alignments. 
However, some datasets may lack granularity structures, posing challenges to granularity constraints. 
Notably, one of the prevailing trends in AI research is the pursuit of large-scale datasets, leading to a significant increase in the number of data categories \cite{recursive}. 
Such large-scale datasets inherently exhibit hierarchical multi-granularity attributes; for instance, the ImageNet dataset utilizes WordNet for granularity structure delineation \cite{imagenetijcv}.
Besides, pioneers have emerged in the research of automatic granularity discovery and construction, such as community discovery \cite{community1}. 
Researchers can upgrade techniques for the automatic construction of granularity based on the commonalities and specificities proposed in this paper.
In addition, considering the distance-based nature of FSDG, the incorporation of
optimal transport-based training objectives is another promising direction for further boosting performance \cite{sk}. This endeavor would help us gain a deeper understanding of the commonalities and specificities among the samples, as well as the gaps between them and their centroids.
Another essential future direction deserving of further exploration is the deeper analysis of the explainability. 
FSDG seeks to render the black-box features of deep learning more transparent, allowing us to identify certain features responsible for handling commonalities and specificities.
Thanks to the numerous breakthroughs in techniques witnessed in recent years, we anticipate a surge of innovation in these promising avenues. 
These further explorations foster the synergistic optimization of feature structuralization.
In summary, our work establishes a strong baseline for exploiting FGDG problems, and we believe that the findings presented in this paper warrant further exploration.


\bibliographystyle{IEEEtran}

\bibliography{MM-021105-FSDG}

\vfill

\end{document}